\documentclass[letterpaper]{article} 
\usepackage{aaai25}  
\usepackage{times}  
\usepackage{helvet}  
\usepackage{courier}  
\usepackage[hyphens]{url}  
\usepackage{graphicx} 
\usepackage{natbib}  
\usepackage{caption} 
\usepackage{algorithm}
\usepackage{algorithmic}
\usepackage{newfloat}
\usepackage{listings}
\DeclareCaptionStyle{ruled}{labelfont=normalfont,labelsep=colon,strut=off} 
\floatstyle{ruled}
\newfloat{listing}{tb}{lst}{}
\floatname{listing}{Listing}
\usepackage{amsmath}
\usepackage{xcolor}
\usepackage{cite}
\usepackage{amssymb}
\usepackage{multirow, booktabs}
\usepackage{svg}
\usepackage{makecell}
\usepackage{subcaption}

\title{Int2Planner: An Intention-based Multi-modal Motion Planner for Integrated Prediction and Planning}
\author{
    Xiaolei Chen\textsuperscript{\rm 1,2},
    Junchi Yan\textsuperscript{\rm 1}\thanks{Corresponding authors.},
    Wenlong Liao\textsuperscript{\rm 2},
    Tao He\textsuperscript{\rm 2,3}\footnotemark[1], 
    Pai Peng\textsuperscript{\rm 2}\thanks{Project Lead.}
}
\affiliations{
    \textsuperscript{\rm 1} School of Artificial Intelligence \&  Department of CSE \& MoE Lab of AI, Shanghai Jiao Tong University\\
    \textsuperscript{\rm 2} COWAROBOT Co. Ltd.\\
    \textsuperscript{\rm 3} School of Electronic Engineering, University of South China\\

    \{cxlz64, yanjunchi\}@sjtu.edu.cn, \{volans.liao, tommie.he\}@cowarobot.com, pengpai\_sh@163.com
}

\usepackage{bibentry}

\begin{document}

\maketitle

\begin{abstract}
Motion planning is a critical module in autonomous driving, with the primary challenge of uncertainty caused by interactions with other participants. As most previous methods treat prediction and planning as separate tasks, it is difficult to model these interactions. Furthermore, since the route path navigates ego vehicles to a predefined destination, it provides relatively stable intentions for ego vehicles and helps constrain uncertainty. On this basis, we construct Int2Planner, an \textbf{Int}ention-based \textbf{Int}egrated motion \textbf{Planner} achieves multi-modal planning and prediction. Instead of static intention points, Int2Planner utilizes route intention points for ego vehicles and generates corresponding planning trajectories for each intention point to facilitate multi-modal planning. The experiments on the private dataset and the public nuPlan benchmark show the effectiveness of route intention points, and Int2Planner achieves state-of-the-art performance. We also deploy it in real-world vehicles and have conducted autonomous driving for hundreds of kilometers in urban areas. It further verifies that Int2Planner can continuously interact with the traffic environment. Code will be avaliable at https://github.com/cxlz/Int2Planner.
\end{abstract}

%

\section{Introduction}
\label{sec:intro}
In autonomous driving, motion planning is a crucial task~\cite{plan1, plan2} to operate ego vehicles without human intervention. Autonomous vehicles are equipped with multiple sensors that continuously observe the environment like a human driver. A key objective of this observation is to assess the movements of the surrounding agents, which is essential to create a motion plan that ensures safe, comfortable, and reliable driving. The trajectories of the surrounding agents have a significant impact on motion planning~\cite{review}.

However, many existing methods traditionally treat motion planning and trajectory prediction as distinct tasks. Typically, these methods first generate predictions of future trajectories for surrounding agents, which are then used as inputs for the planning task. However, they often neglect the interaction between autonomous vehicles and surrounding agents. Therefore, there is a critical need to integrate motion planning and trajectory prediction into a unified model.

In addition, the primary goal of planning is to reach the destination safely by the navigation of the route path, which determines the potential intention of autonomous vehicles. The route path can be integrated into planning models by serving as input to the model and establishing a route-oriented cost function~\cite{sadat2020perceive}. Recent models incorporate attention mechanisms to facilitate the interaction between autonomous vehicles and route paths~\cite{gc-pgp2023prediction, val142023parting}. 

Inspired by the use of target points and endpoints in prediction tasks~\cite{zhao2021tnt, gilles2022gohome, shi2022mtr} to reduce the uncertainty of the multi-modal future, 
we propose Int2Planner, a novel intention-based motion planner for integrated prediction and planning tasks. It utilizes intention points sampled from the route path to handle multi-modal planning, which is inapplicable for prediction tasks, as the route paths of surrounding vehicles are usually unknown. 

\begin{figure*}[tb]
  \centering
\includegraphics[width=0.99\linewidth, trim={0cm, 0cm, 2cm, 0cm}, clip]{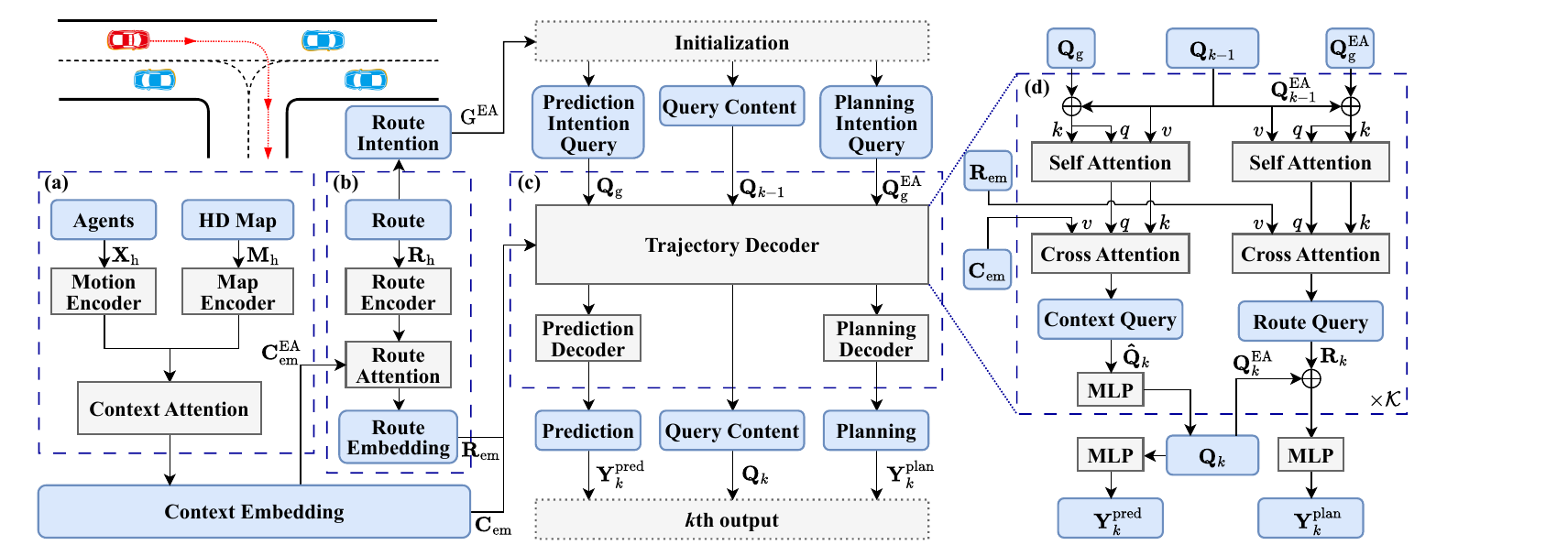}
\caption{The overall framework of Int2Planner. (a) denotes the Context Encoder module, which encodes agent states and HD map information into context embedding, 
(b) denotes the Route Encoder module, which encodes route information into route embedding 
and (c) denotes the Trajectory Generator module, which optimizes future trajectories with $K$ iterations, (d) denotes the detailed structure of the Trajectory Decoder layer and "$\oplus$" denotes a concatenation operation of tensors.
}
  \label{fig:network}
\end{figure*}

Fig.~\ref{fig:network} shows the overall framework of Int2Planner. The main contributions of this paper are listed below:

(1) We develop Int2Planner, a novel planning model, utilizing route intention points to handle the uncertainty of multi-modal planning. It combines prediction and planning in a joint model to realize the interactions between ego vehicles and surrounding agents.

(2) Instead of static intention points, we propose to sample intention points from route path to represent the potential intentions of ego vehicles. Planning trajectories are generated for each intention point to realize multi-modal planning.

(3) We are going to release a new dataset for motion planning tasks. Experiments are conducted on both the private dataset and the public nuPlan dataset. The results demonstrate that the proposed route intention points effectively improve the motion planning ability and Int2Planner achieves state-of-the-art planning performance on these datasets.

(4) We deploy Int2Planner in real-world vehicles, and the test results show that Int2Planner is capable of reacting to complex traffic scenarios and generating safe and reasonable planning trajectories.

\section{Related Work}

\subsubsection{Vehicle Trajectory Prediction}
Early methods~\cite{li2017real, xie2017vehicle} focused primarily on modeling vehicle motion states, resulting in relatively accurate short-term trajectory predictions. 
To address this problem, probabilistic models such as Bayesian networks~\cite{xie2018driving} and decision trees~\cite{hu2017decision} have been used to predict the intentions of agents. Nevertheless, these models can only capture simple close-range interactions between targets.
Trajectory prediction methods based on deep learning have become mainstream, including Convolutional Neural Networks (CNNs)~\cite{ye2021tpcn, gilles2022gohome}, Recurrent Neural Networks (RNNs)~\cite{sun2022m2i, 9349962}, Graph Neural Networks (GNNs) and Attention mechanisms~\cite{9700483, varadarajan2022multipath++}, enabling the consideration of more complex interactions~\cite{li2021attentional}. 
However, information loss in long-time sequences is still a significant issue. Transformers can preserve long-term information from historical trajectories by incorporating position and time embedding, resulting in improvement of performance in trajectory prediction tasks~\cite{nayakanti2023wayformer, ngiam2021scene, jia2023hdgt, zhou2023query}. However, these methods consider trajectory prediction as an independent problem without it with the motion planning task.

\subsubsection{Integrated Motion planning}
Traditional motion planning depends primarily on trajectory clustering methods~\cite{nilsson2013predictive} and optimization-based approaches~\cite{9415170}, frequently resulting in suboptimal planning in a dynamically changing environment. 
Recently, motion planning methods based on machine learning have emerged to account for the evolving environment, enabling the integration of prediction and planning within a joint model. One approach of integrated planning is Robot Leader planning~\cite{schmerling2018multimodal}, where the predictions of surrounding agents are made based on the planning of the ego agent. Consequently, from the viewpoint of the ego agent, the surrounding agents conform to its planning behavior, which can lead to aggressive driving behavior. On the contrary, Human Leader planning methods~\cite{Casas_2021_CVPR} aim to eliminate aggressive driving behavior, but they lack consideration for the impact of the ego agent's planning on the behavior of surrounding agents, resulting in less confident planning. To address these problems, integrated planning should consider the interaction of both ego agent and surrounding agents~\cite{plan1, ye2023fusionad, huang2023gameformer}. Recent work has also explored end-to-end models~\cite{uniad_Hu_2023_CVPR, jiang2023vad} for planning purposes to eliminate the errors introduced by the front-end process and to consider the interaction between prediction and planning.

\subsubsection{Intention in Autonomous Driving}
Intention plays a pivotal role in prediction and planning tasks. 
Early trajectory prediction methods describe multi-modal intentions as action-based~\cite{casas2018intentnet} or region-based~\cite{liu2021multimodal} predictions. 
Recent studies~\cite{rhinehart2019precog, fang2020tpnet} introduce the concept of target points to represent vehicle intentions and generate predicted trajectories based on these target points. 
Some methods~\cite{varadarajan2022multipath++, ngiam2021scene} employ latent anchor features to generate target points and optimize these features through model training. 
Other approaches~\cite{zhao2021tnt} combine map information and sample target points from map lane lines. 
However, since the actual driving intention of a predicted vehicle cannot be known in advance, the sampled target points may not reflect the vehicle's true driving intention. 
DenseTNT~\cite{densetnt} addresses this issue through dense point sampling, but the increase in target points can impact the model's inference speed. MTR~\cite{shi2022mtr} proposes using offline-generated clustering points to describe the formal intentions of different traffic participants. By using clustering methods, target points can be generated based on different driving behaviors, potentially improving adaptability to specific scenarios. 
Although target points have been extensively studied in prediction tasks, there is relatively little research in the field of motion planning. Since the intention of the ego vehicle is often determined by its route path, we suggest that resampling the route points can help eliminate irrelevant target points and thus reduce uncertainty. This approach is inapplicable for prediction tasks~\cite{deo2022multimodal}, as the route paths of surrounding vehicles are usually unknown.

\section{Methodology}
As shown in Fig.~\ref{fig:network}, the overall network of Int2Planner is divided into three main components: Context Encoder, Route Encoder, and Trajectory Generator. Details of each module are introduced in this section. 
\subsection{Preliminaries} \label{sec:input}
Commonly, a typical traffic scenario involves multiple types of vehicles, including Ego Agent (EA) and Surrounding Agents (SAs). The state histories of vehicles are adopted as the primary input features for the prediction and planning tasks. At time step $t_0$, the state history of a typical vehicle in the past $t_\text{h}$ time steps is $X_i=\left\{x_t\mid t\in\left[t_0-t_\text{h},t_0 \right]\right\}$.
Hence, the state histories of all the agents are denoted as:
\begin{eqnarray}
\mathbf{X}_{\text{h}} = \left\{\operatorname{X}^{\text{EA}}, \mathbf{X}^{\text{SA}}\right\} = \left\{X_i\mid i\in\left[0,N_a\right]\right\}
\end{eqnarray}
where $N_a$ is the number of SAs, $\operatorname{X}^{\text{EA}}=X_0$ denotes the state history of EA, and $\mathbf{X}^{\text{SA}}=\left\{X_i\mid i\in\left[1, N_a\right]\right\}$ denotes the state histories of SAs.

In addition, map polylines from HD map are also used as input elements to provide environmental information. A typical map polyline, composed of $t_m$ points, is $P_i = \left\{p_t\mid t\in[1,t_m]\right\}$.
Hence, all the map polylines within a specific range around EA are denoted as:
\begin{eqnarray}
\mathbf{M}_\text{h} = \left\{P_i \mid i\in[1,N_m]\right\}
\end{eqnarray}
where $N_m$ is the number of map polylines.

Furthermore, for planning tasks, route information is also a pivotal input feature for navigating EA to specific destinations. The route information used in our model is divided into two main parts. Firstly, a set of map polylines along the route to the destination is provided as the input features of the Route Encoder module. Similar to map polylines, the route polylines are denoted as:
\begin{eqnarray}
\mathbf{R}_\text{h} = \left\{P_i\mid i\in[1,N_r]\right\}
\label{eq:route_polyline}
\end{eqnarray}
where $N_r$ is the number of route polylines.
Secondly, a set of intention points sampled from these map polylines along the route is adopted as the initialization for EA intention queries in the Trajectory Generator module. These intention points are sampled at an equal distance interval along the route polylines:
\begin{eqnarray}
& \operatorname{G}^{\text{EA}} = \left\{g_i\mid i\in[1,N_q]\right\}
\label{eq:intent_EA}
\end{eqnarray}
where $N_q$ is the number of intention points $g_i$. 
\begin{figure}[t!]
  \centering
\includegraphics[width=0.95\linewidth]{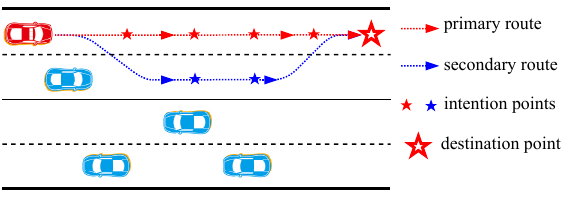}
\caption{Explanation of route intention points sampling.}
\label{fig:route}
\end{figure}

\noindent \textbf{Route Intention Point.} As shown in Fig.~\ref{fig:route}, EA is navigated to a unique destination, while the routes to the destination are multiple and continuously adjusted according to traffic conditions. Among these routes, one optimal route is selected as the primary route, and the others are secondary routes. Intention points of EA can be sampled from the primary route, as well as the secondary routes. These route intention points serve as short-term destinations for EA and corresponding planning trajectories are generated for each intention point. The impact of different sampling strategies is investigated in the experimental section.

For both prediction and planning tasks, the output features are represented as trajectories and each trajectory consists of ${t_\text{f}}$ future points:
\begin{eqnarray}
    & \mathbf{Y}^{\text{pred}} &= \left\{Y_i^{\text{pred}}\mid i\in[0,N_a]\right\}\\
    & \operatorname{Y}^{\text{plan}} &= Y_0^{\text{pred}}
\end{eqnarray}
where $Y_i^{\text{pred}} = \left\{Y_t\mid t\in[t_0+1,t_0+t_\text{f}]\right\}$.

\subsection{Context Encoder}
The Context Encoder module integrates agents and HD map information into context embedding. Firstly, $\mathbf{X}_{\text{h}}$ and $\mathbf{M}_{\text{h}}$ are adopted as input features, and both are processed by a PointNet-like encoder, which generally consists of a Multilayer Perceptron (MLP) layer followed by a maxpooling operation, denoted as $\mathbf{\phi}_{\text{agg}}\left(*\right) = \mathbf{Maxpool}\left(\mathbf{MLP}\left(*\right)\right)$. Then the aggregated agent and map embeddings are concatenated and go through a Context Attention module, which is implemented as a Self-Attention layer.
\begin{eqnarray}
& \mathbf{C}_{\text{em}} = \mathbf{\phi_{\text{context}}}\left(\mathbf{\phi}_{\text{agg}}\left(\left[\mathbf{X}_{\text{h}}, \mathbf{M}_{\text{h}}\right]\right)\right)
\label{eq:scene_embedding}
\end{eqnarray}
where $\mathbf{\phi_{\text{context}}}\left(*\right) = \mathbf{SelfAttn}\left(*\right)$.
Through this attention module, all context elements can interact with each other.
 
\subsection{Route Encoder}
Similar to the Context Encoder module, the Route Encoder takes route polylines as input, and then aggregates route feature $\mathbf{R}_{\text{agg}}$ by a PointNet-like encoder. Since only EA follows the guidance of route polylines, the context embedding of EA, denoted as $\operatorname{C}_{\text{em}}^{\text{EA}}$, is separated from the entire context embedding, which then goes through a Route Attention module to generate route embedding. Route embedding $\mathbf{R}_{\text{em}}$ is denoted as:
\begin{eqnarray}
& \mathbf{R}_{\text{em}} &= \mathbf{\phi_{\text{route}}}\left(\left[\operatorname{C}_{\text{em}}^{\text{EA}}, \mathbf{\phi}_{\text{agg}}\left(\mathbf{R}_{\text{h}}\right)\right]\right)
\label{eq:route_embedding}
\end{eqnarray}
where $\mathbf{\phi_{\text{route}}}\left(*\right) = \mathbf{SelfAttn}\left(*\right)$ is also a Self-Attention layer. Hence, only EA interacts with route polylines.
\subsection{Trajectory Generator}
Trajectory Generator mainly contains a Transformer-based decoder that processes the input context embedding and route embedding. The prediction trajectory $\mathbf{Y}^{\text{pred}}$ and planning trajectory $\operatorname{Y}^{\text{plan}}$ are refined through an iteration process with $K$ iterations. 

To maintain the decoded features during the iteration process, the query content $\mathbf{Q}_{k}$ is utilized and updated every iteration. 
In the first iteration, the initial query content $\mathbf{Q}_{1}$ is initialized by a trainable embedding tensor. 

Following~\cite{shi2022mtr}, we utilize static intention points to initialize intention queries and a corresponding trajectory is generated for each intention point.
Since EA is navigated with a predefined destination, instead of utilizing clustered points, the intention points of EA are sampled from route polylines $\mathbf{R}_\text{h}$ with an equal distance interval $d_r$, as illustrated in Eq.~(\ref{eq:intent_EA}). The intention points of all the agents are denoted as:
\begin{eqnarray}
    \mathbf{G} = \left\{\operatorname{G}^{\text{EA}}, \mathbf{G}^{\text{SA}}\right\} = \left\{G_i\mid i\in[0,N_a]\right\}
\end{eqnarray}
where $\operatorname{G}^{\text{EA}}=G_0$ is the intention points of EA and $\mathbf{G}^{\text{SA}}=\{G_i\mid i\in[1,N_a]\}$ is the intention points of SA. The Intention Query are initialized as:
\begin{eqnarray}
    \mathbf{Q}_{\text{g}} = \mathbf{\phi}_\text{g}\left(\mathbf{G}\right)
\end{eqnarray}
where $\mathbf{\phi}_\text{g}$ is implemented as an MLP layer.

The details of the Trajectory Decoder layer is shown in the right part of Fig.~\ref{fig:network}, which basically consists of several transformer decoder layers. In the decoder part of the prediction task, query content $Q_{k-1}$ from previous iteration firstly goes through a Self-Attention layer, with the intention query $\mathbf{Q}_{\text{g}}$ as position embedding. Secondly, the output of the Self-Attention layer is used as both the key and query of the Cross-Attention layer to extract corresponding values from context embedding $\mathbf{C}_{\text{em}}$, updating the query content to $\mathbf{Q}_{k}$. Finally, the updated query content is processed through an MLP layer to generate the prediction trajectory points as the outputs of $k$th iteration.
\begin{eqnarray}
& \mathbf{Q}_{k} = \mathbf{\phi}_{\text{tr}}\left(\mathbf{Q}_{k-1}, \mathbf{Q}_{\text{g}},\mathbf{C}_{\text{em}}\right) 
\label{eq:tr} \\
& \left(\mathbf{Y}^{\text{pred}}_{k}, \mathbf{S}_{k}^{\text{pred}}\right) = \mathbf{\phi}_\text{f}\left(\mathbf{Q}_{k}\right) 
\label{eq:detr}
\end{eqnarray}
where $\mathbf{\phi}_\text{f}\left(*\right)=\mathbf{MLP}\left(*\right)$ is an MLP layer, $\mathbf{\phi}_{\text{tr}}\left(*\right)=\mathbf{CrossAttn}\left(\mathbf{SelfAttn}\left(*\right)\right)$, $\mathbf{SelfAttn}$ and $\mathbf{CrossAttn}$ are Self-Attention layer and Cross-Attention layer, respectively. $\mathbf{S}_{k}^{\text{pred}}$ is the confidence score of multi-modal prediction trajectories.

\begin{table*}[t!]
\begin{center}
\begin{tabular}{@{} lcccc|cccc @{} }
\toprule 
\multirow{2}{*}{Models} & \multicolumn{4}{c|}{Val14} & \multicolumn{4}{c}{Test14-hard} \\
\cmidrule{2-9}
& Overall~($\uparrow$) & OL~($\uparrow$) & NR-CL~($\uparrow$) & R-CL~($\uparrow$) & Overall~($\uparrow$) & OL~($\uparrow$) & NR-CL~($\uparrow$) & R-CL~($\uparrow$) \\
\midrule 
\textit{Log-replay}  & \textit{0.9133}  & \textit{1.00} & \textit{0.94} & \textit{0.80} & \textit{0.8500} & \textit{1.00} & \textit{0.86} & \textit{0.69} \\
\midrule
IDM & 0.6367  &  0.38 & 0.76 & \textbf{0.77} & 0.4600 &  0.20 & 0.56 & \underline{0.62} \\
PlanCNN  & 0.6967  & 0.64 & 0.73 & 0.72 & - & - & - & - \\
GC-PGP & 0.6433  & 0.82 & 0.57 & 0.54 & 0.5221 & 0.7378 & 0.4322 & 0.3963 \\
PlanTF & \textbf{0.8360} & \underline{0.8918} & \textbf{0.8483} & \textbf{0.7678} & \underline{0.7263} & \underline{0.8332} & \textbf{0.7286} & 0.6170 \\
Int2Planner (ours) & \underline{0.8226} & \textbf{0.9097} & \underline{0.7912} & \underline{0.7668} & \textbf{0.7476} & \textbf{0.8673} & \underline{0.6971} & \textbf{0.6784} \\
\midrule 
GameFormer Planner~*  & 0.8216  & 0.8304 & 0.8182 & {0.8161} & 0.7035 & \underline{0.7527} & \underline{0.6695} & {0.6883} \\
PDM-Hybrid~*  & \textbf{0.8967} & \textbf{}0.84 & \textbf{0.93} & \textbf{0.92} & \underline{0.7185} & 0.7381 & {0.6595} & \textbf{0.7579} \\
Int2Planner~* (ours) & \underline{0.8385} & \textbf{0.8513} & \underline{0.8372} & \underline{0.8269} & \textbf{0.7679} & \textbf{0.8079} & \textbf{0.7500} & \underline{0.7457} \\
\bottomrule 
\end{tabular}
\caption{Simulation results on nuPlan benchmark. 
"*" indicates the hybrid models combine learning-based and rule-based methods.
"OL", "NR-CL" and "R-CL" indicate open-loop,  non-reative closed-loop and reactive closed-loop simulations, respectively. 
"Overall" indicates the average score of three simulations.} 
\label{tab:simulation}
\end{center}
\end{table*}

On the other hand, the planning part utilizes a similar flow. What makes it different are three points. One is that only the query content of EA $\operatorname{Q}_{k-1}^{\text{EA}}$ and the planning intention query $\operatorname{Q}_{\text{g}}^{\text{EA}}$ are used in the Self-Attention layer. The second difference lies in the Cross-Attention layer, in which the route embedding $\mathbf{R}_{\text{em}}$ is used as the value to generate the route content $\mathbf{R}_{k}$. The last is that the updated query content $\operatorname{Q}_{k}^{\text{EA}}$ of EA is concatenated with route content $\mathbf{R}_{k}$ to generate the planning trajectory points of EA, considering both context and route features.
\begin{eqnarray}
& \mathbf{R}_{k} = \mathbf{\phi}_{\text{tr}}\left(\operatorname{Q}_{k}^{\text{EA}}, \operatorname{Q}_{\text{g}}^{\text{EA}},\mathbf{R}_{\text{em}}\right) \\
& \left(\operatorname{Y}^{\text{plan}}_{k}, \operatorname{S}_{k}^{\text{plan}}\right) = \mathbf{\phi}_\text{f}\left(\left[\operatorname{Q}_{k}^{\text{EA}}, \mathbf{R}_{k}\right]\right) 
\label{eq:gmm}
\end{eqnarray}
where $\operatorname{S}_{k}^{\text{plan}}$ is the confidence score of multi-modal planning trajectories.

The results of the $K$th iteration are utilized as the final output of the Trajectory Generator. 

\subsection{Loss Function}
The training process of Int2Planner is supervised by GT trajectories, with L1 loss for trajectory regression and cross-entropy loss for confidence score. Following~\cite{shi2022mtr}, the intention point closest to the endpoint of GT trajectory is selected as the positive item, which is used to calculate the confidence score loss. The predicted trajectory corresponding to the selected intention point is used to calculate the trajectory regression loss. 
In addition, the final loss is the mean value of all the losses from $K$ iterations.

\section{Experiments}
\label{sec:experiments}
\subsection{Experiments Setup}
\subsubsection{Datasets and Metrics}
We conduct experiments on a private dataset from an autonomous driving corporation. This dataset includes extensive trajectory data, localization data and route path information, making it suitable for both prediction and planning tasks. The data, collected primarily in urban environments, comes from autonomous vehicles operating under various conditions, such as daytime vs. night and sunny vs. rainy weather. These vehicles are equipped with a central roof-mounted sensor unit, which consists of one Ruby Plus 128\footnote{\url{https://www.robosense.ai/en/rslidar/RS-Ruby_Plus}} Lidar and five standard cameras for panorama vision. In addition, four fisheye cameras are installed around the vehicles for close-range vision. 
To generate valid ground-truth planning trajectories, all data is collected through manual driving by expert vehicle operators. Each scene contains an average of about 43 agents, detected and tracked by a state-of-the-art offline perception system. 

The private dataset contains 680,964 traffic scenarios and each scenario contains 6.5 seconds of trajectory data at 10 Hz,
in which 626,459 scenarios are used as train set and the rest 54,505 scenarios remain as validation set. Following the common usage, ADE and FDE are used as evaluation metrics for prediction and planning tasks. This dataset will be available at https://github.com/cxlz/Int2Planner.

In addition, experiments are also conducted on nuPlan~\cite{plancnn2021nuplan}, a large-scale benchmark for planning tasks in autonomous driving. 
It provides a closed-loop simulator for three simulation tasks: open-loop (OL) planning, nonreactive closed-loop (NR-CL) planning and reactive closed-loop (R-CL) planning. For each task, a weighted score is calculated considering various metrics. 
As the online simulation engine for the nuPlan test set is closed, we use the Val14~\cite{val142023parting} and Test14-hard~\cite{plantf2023} benchmarks for evaluation.

\subsubsection{Implementation Details}
We train Int2Planner on 8 NVIDIA RTX 4090 GPUs for 30 epochs with a total batch size of 96. During the training process, AdamW optimizer is used with an initial learning rate of $1 \times 10^{-4}$ and a weight decay of 0.01. For the private dataset, we use 
$t_\text{h}=15$ historical points and $t_\text{f}=50$ future points for each agent. The feature dimension of context embedding, route embedding and the hidden dimension of attention layers are all set to 128. The distance interval $d_r$ to sample route intention points is set to 4 meters, the number of intention points $N_q$ is set to 64 and the number of decoder iterations $K$ is set to 6.
The range of map polylines is approximately 200 meters. For the nuPlan dataset, we adjust $t_\text{h}=20$ and $t_\text{f}=80$ to match the requirements of nuPlan benchmark, while keeping other hyper-parameters unchanged.
\begin{figure}[t!]
  \centering
\begin{subfigure}{0.48\linewidth}
  \centering
    \includegraphics[width=1\linewidth]{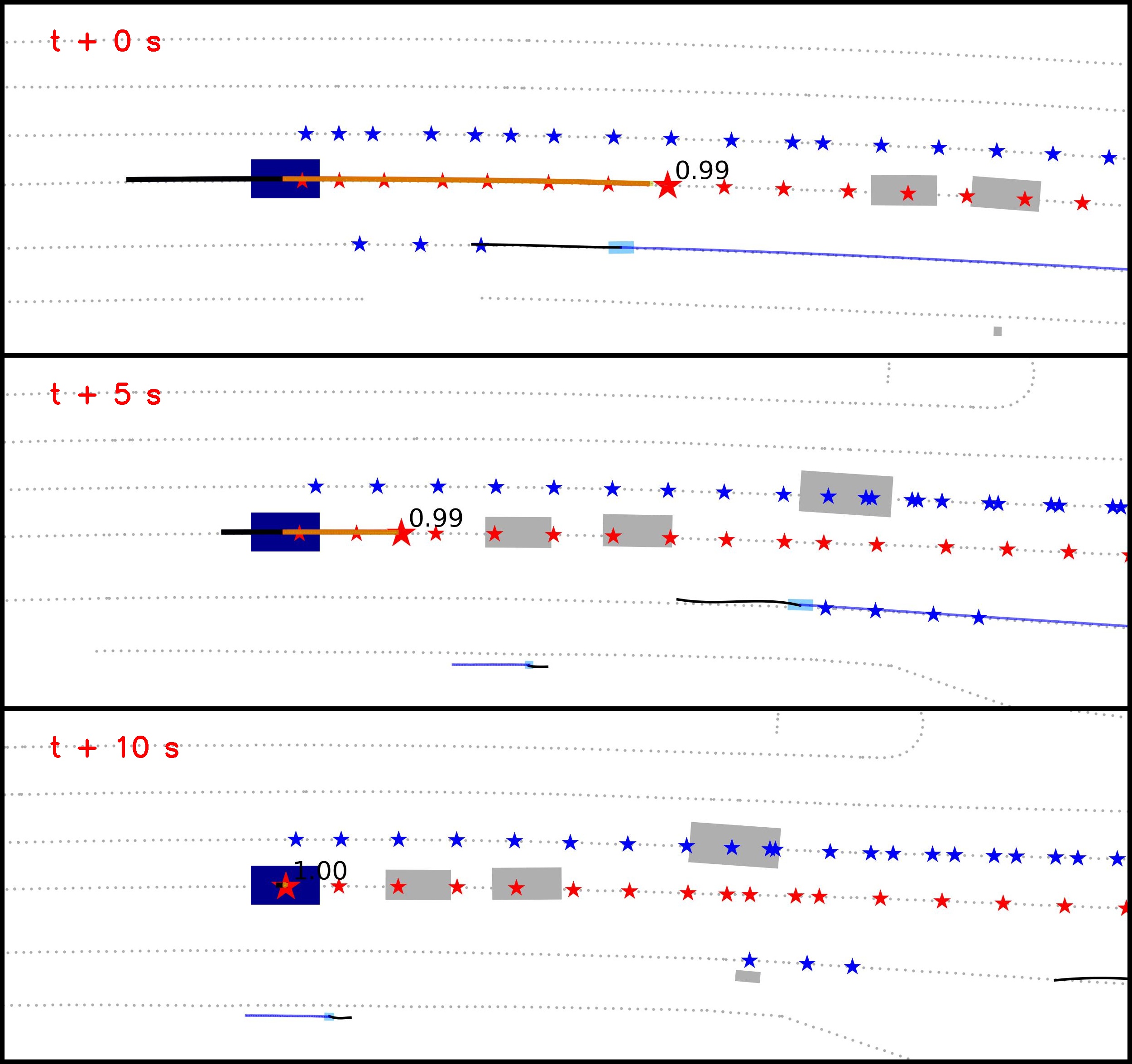}
    \caption{Stop behind agent}
    \label{fig:vis-a}
\end{subfigure}
\begin{subfigure}{0.48\linewidth}
  \centering
    \includegraphics[width=1\linewidth]{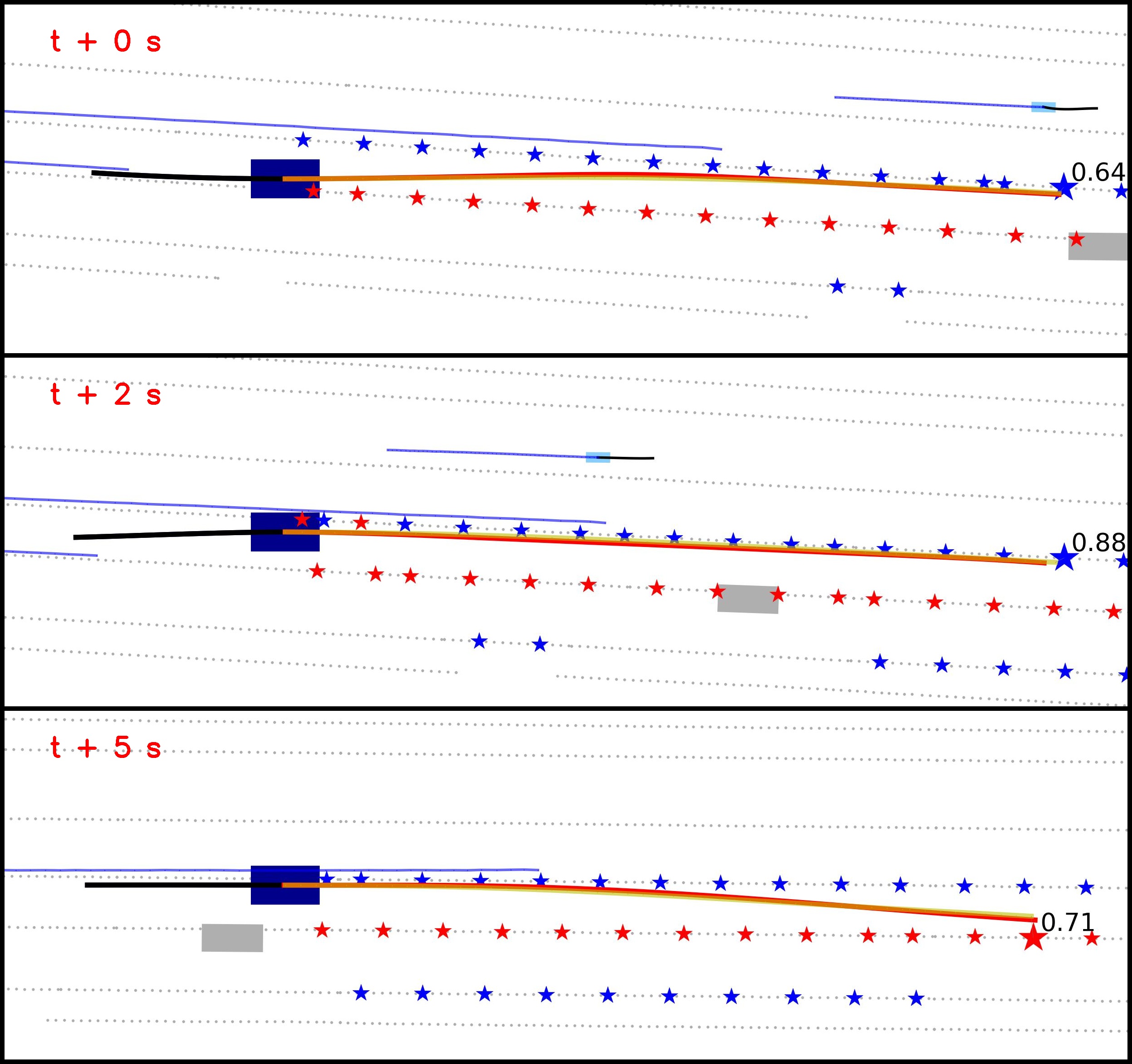}
    \caption{Avoid parked agent}
    \label{fig:vis-b}
\end{subfigure} \\
\centering
\begin{subfigure}{0.48\linewidth}
  \centering
    \includegraphics[width=1\linewidth]{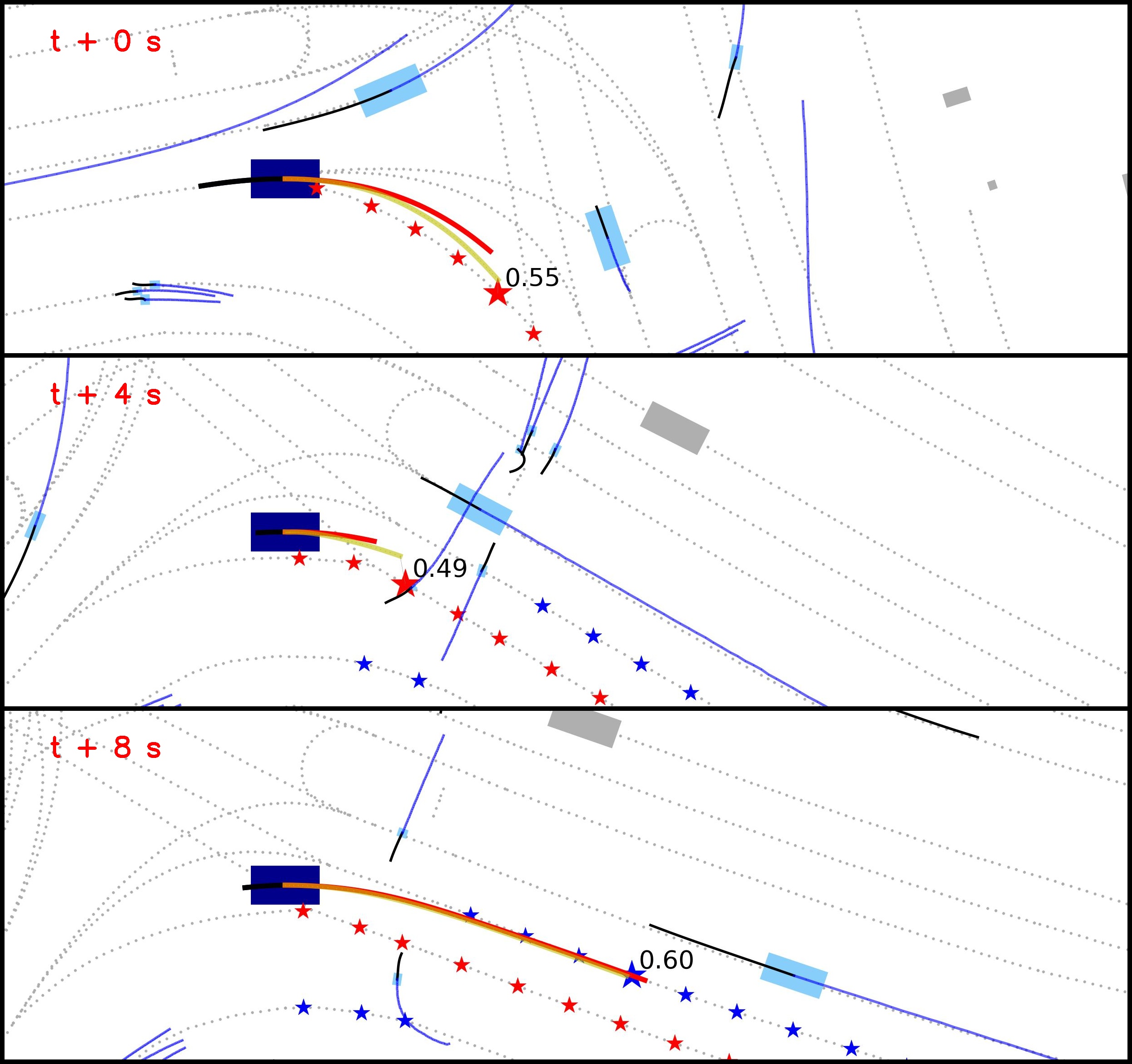}
    \caption{Turn right}
    \label{fig:vis-c}
\end{subfigure}
\begin{subfigure}{0.48\linewidth}
  \centering
    \includegraphics[width=1\linewidth]{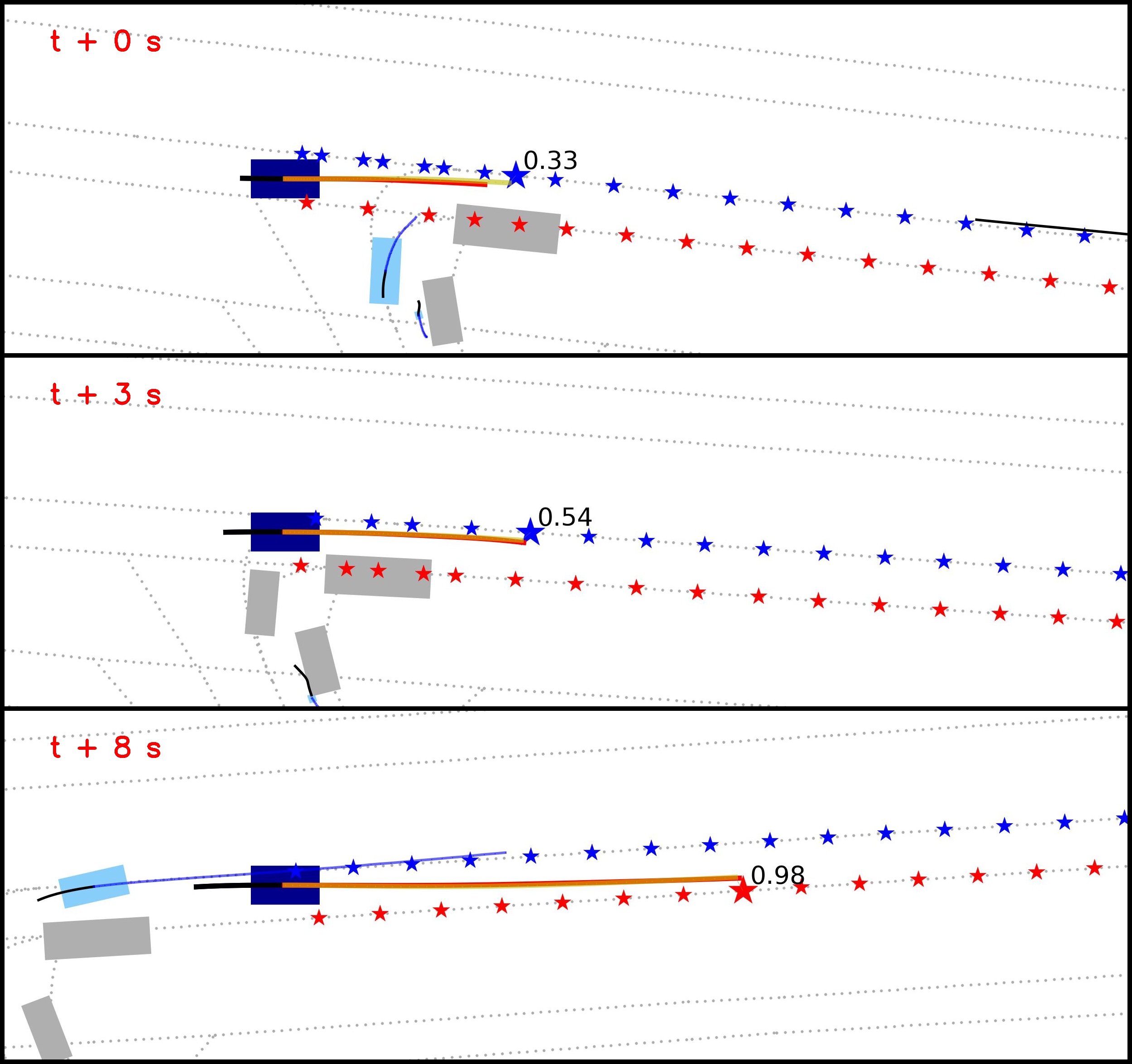}
    \caption{Interaction with agents}
    \label{fig:vis-d}
\end{subfigure} \\
\centering
\begin{subfigure}{0.28\linewidth}
  \centering
    \includegraphics[height=0.88cm]{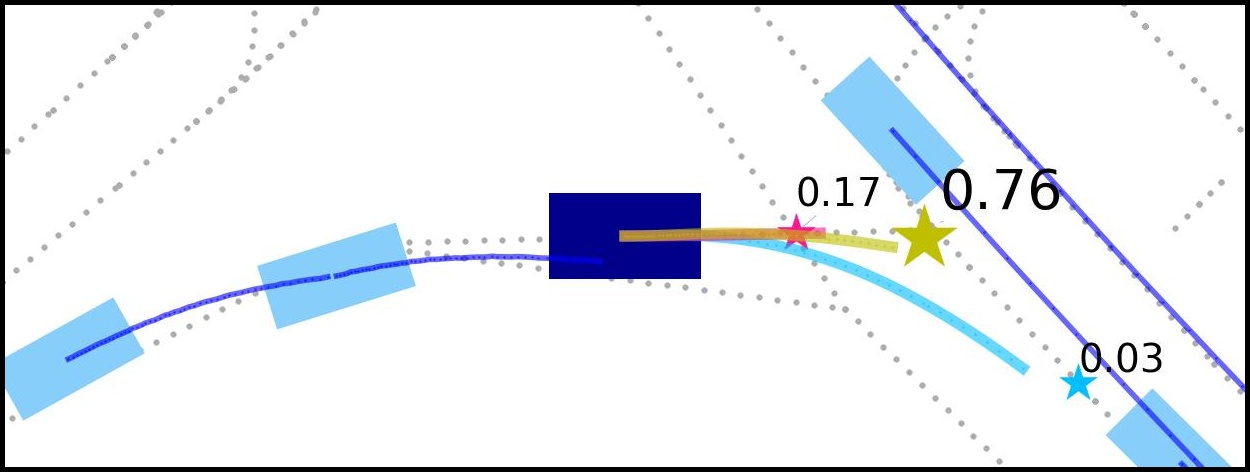}
    \caption{}
    \label{fig:multi-a}
\end{subfigure}
\begin{subfigure}{0.28\linewidth}
  \centering
    \includegraphics[height=0.88cm]{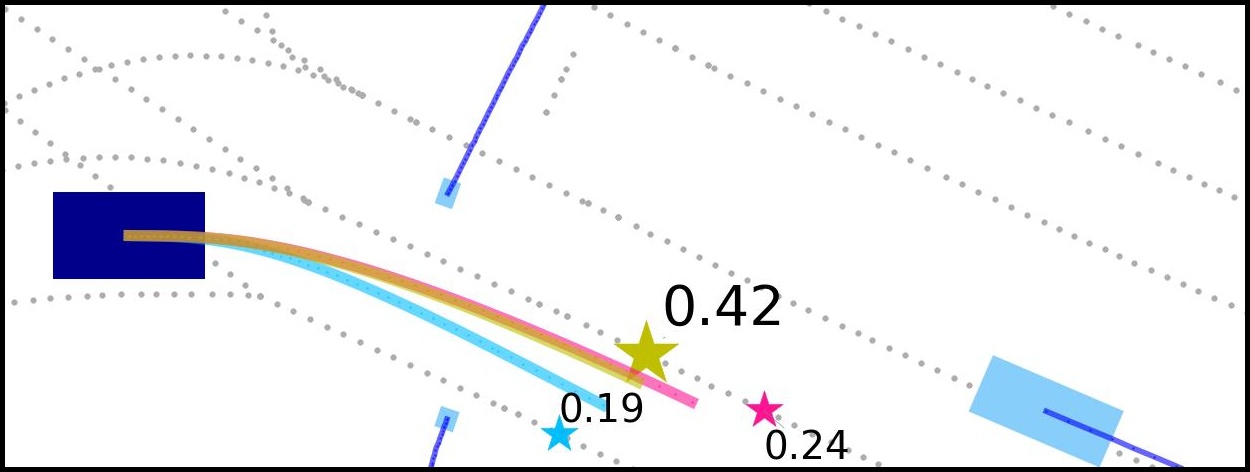}
    \caption{}
    \label{fig:multi-b}
\end{subfigure}
\begin{subfigure}{0.4\linewidth}
  \centering
    \includegraphics[height=0.88cm]{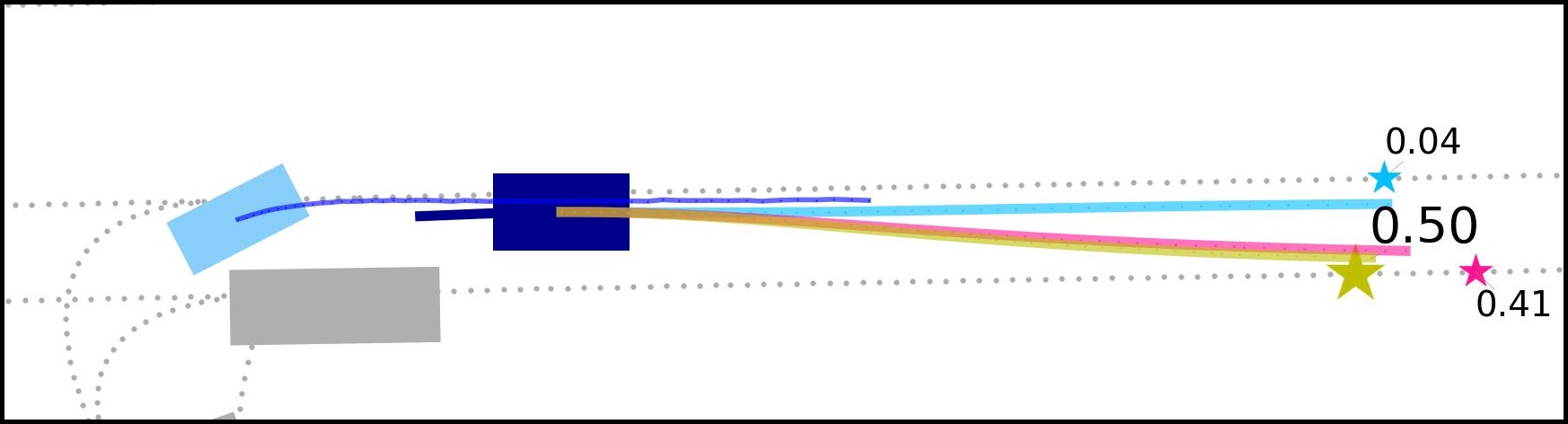}
    \caption{}
    \label{fig:multi-c}
\end{subfigure}
\caption{Qualitative results. (a)-(d) Primary and secondary intention points are marked with red and blue "$\star$". Intention points with the highest confidence are marked larger, along with their confidence scores. (e)-(g)
The three most confident planning trajectories are plotted with intention points
and confidence scores.
}
\label{fig:visualization}
\end{figure}

\begin{table*}[t!]
\begin{center}
\begin{tabular}{@{} l|lcccc @{} }
\toprule
\multicolumn{2}{l}{\multirow{2}{*}{Model}} & \multicolumn{2}{c}{Plan} & \multicolumn{2}{c}{Prediction} \\
\cmidrule{3-6}
\multicolumn{2}{c}{} & $\operatorname{ADE~}(\downarrow)$ & $\operatorname{FDE~}(\downarrow)$ & $\operatorname{minADE_6~}(\downarrow)$ & $\operatorname{minFDE_6~}(\downarrow)$ \\
\midrule
\multicolumn{2}{l}{GameFormer Planner~\cite{huang2023gameformer}} & 0.5400 & 1.1524 & 1.3957 & 2.0861\\
\multicolumn{2}{l}{Int2Planner (ours, w/o Pred.)} & 0.4028 & {1.1088} & - & - \\   
\midrule
\multirow{3}{*}{\makecell{\\Int2Planner\\(ours)}}
& Cluster Intention           & 0.4788 & 1.3436 & \textbf{0.5654} & \textbf{1.1592}  \\
& Route Intention (Primary)& {0.4141} & {1.1505} & 0.6067 & 1.3049 \\         
& Route Intention (All)    & \textbf{0.3946} & \textbf{1.0877} & {0.6010} & {1.2794} \\  
\bottomrule
\end{tabular}
\caption{Evaluations on the private dataset. 
"Cluster Intention" and "Route Intention" indicate that intention points are generated by K-means clustering method and sampled from route polylines, respectively. "Primary" indicates that only the primary route polylines are sampled and "All" indicates that the secondary route polylines are also sampled. 
} 
\label{tab:private}
\end{center}
\end{table*}

\subsection{Main Results}
\subsubsection{Simulation Results}
Table~\ref{tab:simulation} shows the comparison of Int2Planner with other learning-based and rule-based planners, including IDM~\cite{idm2000congested}, PlanCNN~\cite{renz2022plant}, GC-PGP~\cite{gc-pgp2023prediction}, PlanTF~\cite{plantf2023}, PDM-Hybrid~\cite{val142023parting} and GameFormer Planner~\cite{huang2023gameformer}. We train GameFormer Planner and Int2Planner on nuPlan train set and conduct simulations on Val14 and Test14-hard benchmarks. The results of other models are taken from~\cite{val142023parting, plantf2023}.

The simulation results show that Int2Planner reaches state-of-the-art performance overall. Among purely learning-based methods, it demonstrates competitive performance with PlanTF on the Val14 benchmark. For the more complex Test14-hard benchmark, Int2Planner attains the best OL and R-CL simulation scores, as well as the highest overall score. When combined with rule-based post-processing, the performance of Int2Planner on closed-loop simulations is further enhanced. On the Test14-hard benchmark, the simulation scores of Int2Planner are either higher than or close to those of PDM-Hybrid, with an overall score of 0.7679, surpassing PDM-Hybrid by 7\%.

\subsubsection{Evaluation Results}
Table~\ref{tab:private} shows the evaluation results of the prediction and planning performance on the private dataset. When compared to the baseline method, GameFormer Planner, Int2Planner performs better on motion planning metrics, and also shows significant improvement in trajectory prediction performance. Both joint prediction and route intention improve the metrics, while the contributions are different. Although, the model without prediction task does not produce explicit prediction, it still takes SA features as inputs, allowing it to extract SA embeddings for EA planning.
\subsubsection{Qualitative Results}  
Fig.~\ref{fig:vis-a}-\ref{fig:vis-d} shows several common traffic scenarios sampled from the private validation dataset. Bounding boxes of EA, moving and static SAs are colored in deep blue, light blue and grey. Ground-truth, planning, and prediction trajectories are shown in red, yellow and blue. 
The planning trajectory with the highest confidence is highlighted in each figure. 
These scenarios show that Int2Planner interacts with SAs based on the perception and prediction results, making reasonable planning actions.
\subsubsection{Intention Visualization}  
As shown in Fig.~\ref{fig:vis-a}-\ref{fig:vis-d}, the specific intention point corresponding to the planning trajectory with the highest confidence is marked with a larger size and a confidence score. The selected point is almost the closest intention point to the end point of the plotted planning trajectory. This demonstrates that the proposed route intention points offer sufficient potential options and Int2Planner can select appropriate intention points for EA. 

Fig.~\ref{fig:multi-a}-\ref{fig:multi-c} shows the multi-modal intentions in more detail.
The end points of three most confident planning trajectories are distributed relatively sparsely and close to the intention points. It indicates that Int2Planner treats these intention points as short-term destinations, and generates corresponding planning trajectories, which are refined based on the selected intention points.
\subsubsection{Confidence Scores Distribution}
\begin{figure}[t]
  \centering
  \includegraphics[width=0.95\linewidth, trim={0cm, 0cm, 0cm, 0.87cm}, clip]{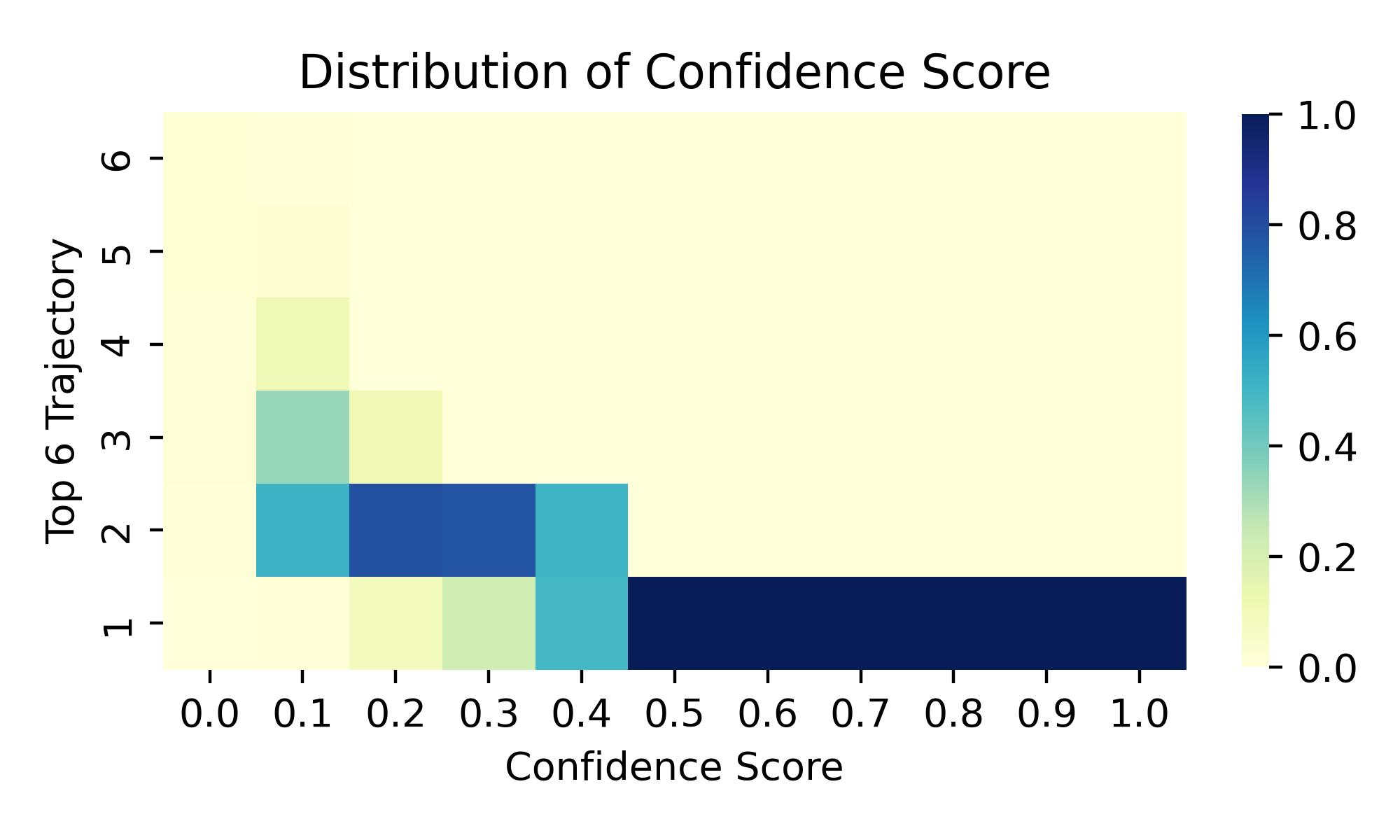}
   \caption{The distribution of planning confidence score.}
   \label{fig:confidence}
\end{figure}

In addition, the confidence score distribution for the planning task for the top 6 intention points are shown in Fig.~\ref{fig:confidence} and the confidence scores tend to concentrate on the top few points. 
\begin{table}[t!]
\centering
\begin{center}
\begin{tabular}{@{} cccccc  @{}}
\toprule
RE & CI & RI & OL~($\uparrow$) & NR-CL~($\uparrow$) & R-CL~($\uparrow$) \\ 
\midrule 
$\times$ & \checkmark & $\times$ & 0.8377 & 0.6775 & 0.6559 \\
$\times$ & $\times$ & \checkmark & 0.8606 & 0.6587 & 0.6760 \\
\checkmark & \checkmark & $\times$ & 0.8453 & 0.6256 & 0.6523 \\
\checkmark & $\times$ & \checkmark & \textbf{0.8673} & \textbf{0.6971} & \textbf{0.6784} \\
\bottomrule 
\end{tabular}
\caption{Effects of route information on Test14-hard benchmark. 
"RE" indicates route embedding, "CI" indicates Cluster Intention, and "RI" indicates Route Intention.
} 
\label{tab:route}
\end{center}
\end{table}
\begin{table}[t!]
\centering
\begin{center}
\begin{tabular}{@{} cccc  @{}}
\toprule
Integrated Prediction& OL~($\uparrow$) & NR-CL~($\uparrow$) & R-CL~($\uparrow$) \\ 
\midrule 
$\times$ & {0.8649} & 0.6784 & 0.6736 \\
\checkmark & \textbf{0.8673} & \textbf{0.6971} & \textbf{0.6784} \\
\bottomrule 
\end{tabular}
\caption{Effects of integrated prediction in Int2Planner.
} 
\label{tab:pred}
\end{center}
\end{table}
\begin{table}[t!]
\centering
\begin{center}
\begin{tabular}{@{} cccc  @{}}
\toprule
Number of Iterations& OL~($\uparrow$) & NR-CL~($\uparrow$) & R-CL~($\uparrow$) \\ 
\midrule
$K$=1& 0.8413 & 0.6154 & 0.6062 \\
$K$=2& 0.8602 & 0.6503 & 0.6416 \\
$K$=3& 0.8622 & 0.6778 & 0.6756 \\
$K$=6& \textbf{}0.8673 & \textbf{0.6971} & \textbf{0.6784} \\
$K$=9& \textbf{0.8736} & \textbf{}0.6871 & \textbf{}0.6764 \\
\bottomrule
\end{tabular}
\caption{Effects of number of decoder iterations.} 
\label{tab:iter}
\end{center}
\end{table}

\begin{table}[t!]
\centering
\begin{center}
\begin{tabular}{@{} cccc  @{}}
\toprule
Outputof $k$th Layer& OL~($\uparrow$) & NR-CL~($\uparrow$) & R-CL~($\uparrow$) \\ 
\midrule
$k$=1& 0.8521 & 0.5651 & 0.6112 \\
$k$=2& 0.8588 & 0.6350 & 0.6504 \\
$k$=3& 0.8606 & 0.6767 & 0.6638 \\
$k$=6& \textbf{0.8673} & \textbf{0.6971} & \textbf{0.6784} \\
\bottomrule
\end{tabular}
\caption{Effects of iteration strategy. The output of $k$th decoder layer is used as final output.} 
\label{tab:layer}
\end{center}
\end{table}

\subsection{Ablation Study}
We conduct ablation studies on the designs of Int2Planner and all the simulation experiments are conducted on Test14-hard benchmark without combining rule-based post-processing.
\subsubsection{Effects of Route Information}
Table~\ref{tab:route} shows the effects of route embedding and Route Intention Points. Route embedding indicates $\mathbf{R}_{\text{em}}$ expressed in Eq.~\ref{eq:route_embedding}. 
The strategies of generating intention points for EA are compared. Cluster intention points are generated by applying the K-means Clustering algorithm on all the endpoints of GT trajectories~\cite{shi2022mtr} and route intention points are sampled from the route path as shown in Fig~\ref{fig:route}. Table~\ref{tab:route} shows that it achieves the best simulation scores, combined with route embedding and Route Intention Points. 

The experiment results on the private dataset shown in Table~\ref{tab:private} further verify that route intention points significantly outperform cluster intention points. In addition,  sampling points from both the primary and secondary route polylines further reduces the planning distance error.
In detail, Fig.~\ref{fig:vis-b} and \ref{fig:vis-d} show that route intention points sampled from secondary route polylines provide EA with additional routing options, especially when the primary route is blocked.  
\subsubsection{Effects of Integrated Prediction}
As shown in Table~\ref{tab:pred}, experiments are conducted on Int2Planner by removing prediction task, to show the effects of integrated prediction. The results show that the planning performance deteriorates without the prediction task, which indicates that combining prediction and planning is necessary.
\subsubsection{Effects of Decoder Iteration}
The value of $K$ is the total number of decoder iterations. Table \ref{tab:iter} shows the performance improvement with the increased number of iterations from 1 to 6. When further increasing the number of iterations to 9, the simulation results does not indicate obvious improvement. Therefore, we finally choose $K=6$ for Int2Planner in the remaining experiments. Table~\ref{tab:layer} demonstrates the performance of the planning trajectories from the $k$th iteration (the $K$ iteration is used by default). The results further verify the effectiveness of the iteration strategy.

\subsection{Real-world Vehicle Test}
We deploy Int2Planner in real-world autonomous driving vehicles. The test vehicles are the same type as those used to collect the private dataset. The perception and track results based on onboard sensors are utilized as input features of Int2Planner to extract historical states of SA. In addition, the route path is generated by a rule-based routing planner. The planning trajectory with the highest confidence of the multi-modal outputs of Int2Planner is passed to the control system to operate the autonomous driving vehicle. 

We have conducted autonomous driving for hundreds of kilometers in urban areas in various scenes. During test experiments, each vehicle is equipped with onboard vehicle operators and a remote monitor system to prevent dangerous driving behaviors that may be caused by model failure.
Fig.~\ref{fig:test} shows several real-world test scenarios. The front view image and four surrounding view images are combined to show the entire environment, and for clarity, the planning trajectory of EA is projected onto the front view image. Each scenario is shown with three sub-images, with relative times noted in the top-left corner.
\begin{figure}[t!]
  \centering
\begin{subfigure}{0.48\linewidth}
  \centering
    \includegraphics[width=1\linewidth]{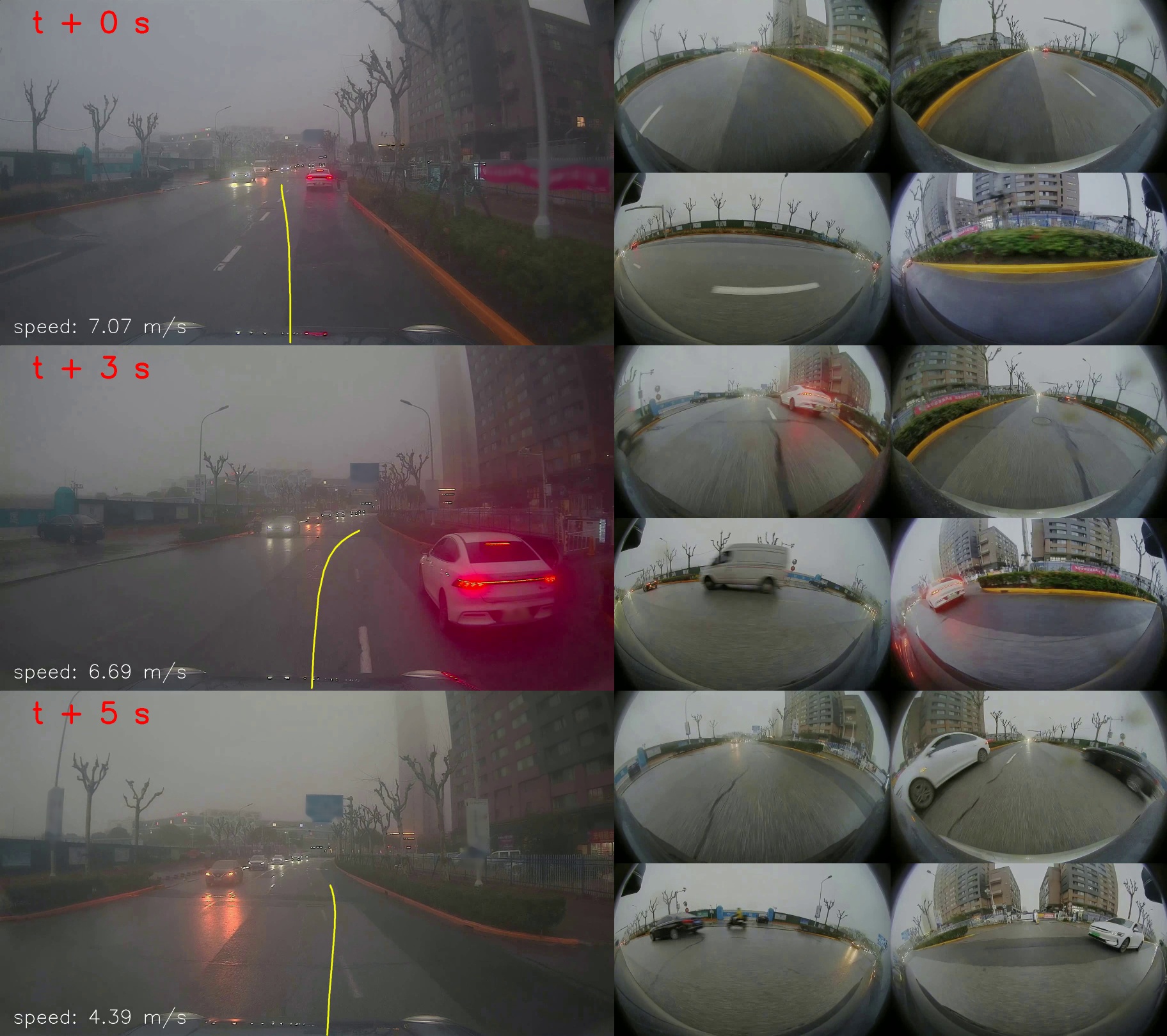}
    \caption{Avoid a vehicle picking up}
    \label{fig:test-A}
\end{subfigure}
\begin{subfigure}{0.48\linewidth}
  \centering
    \includegraphics[width=1\linewidth]{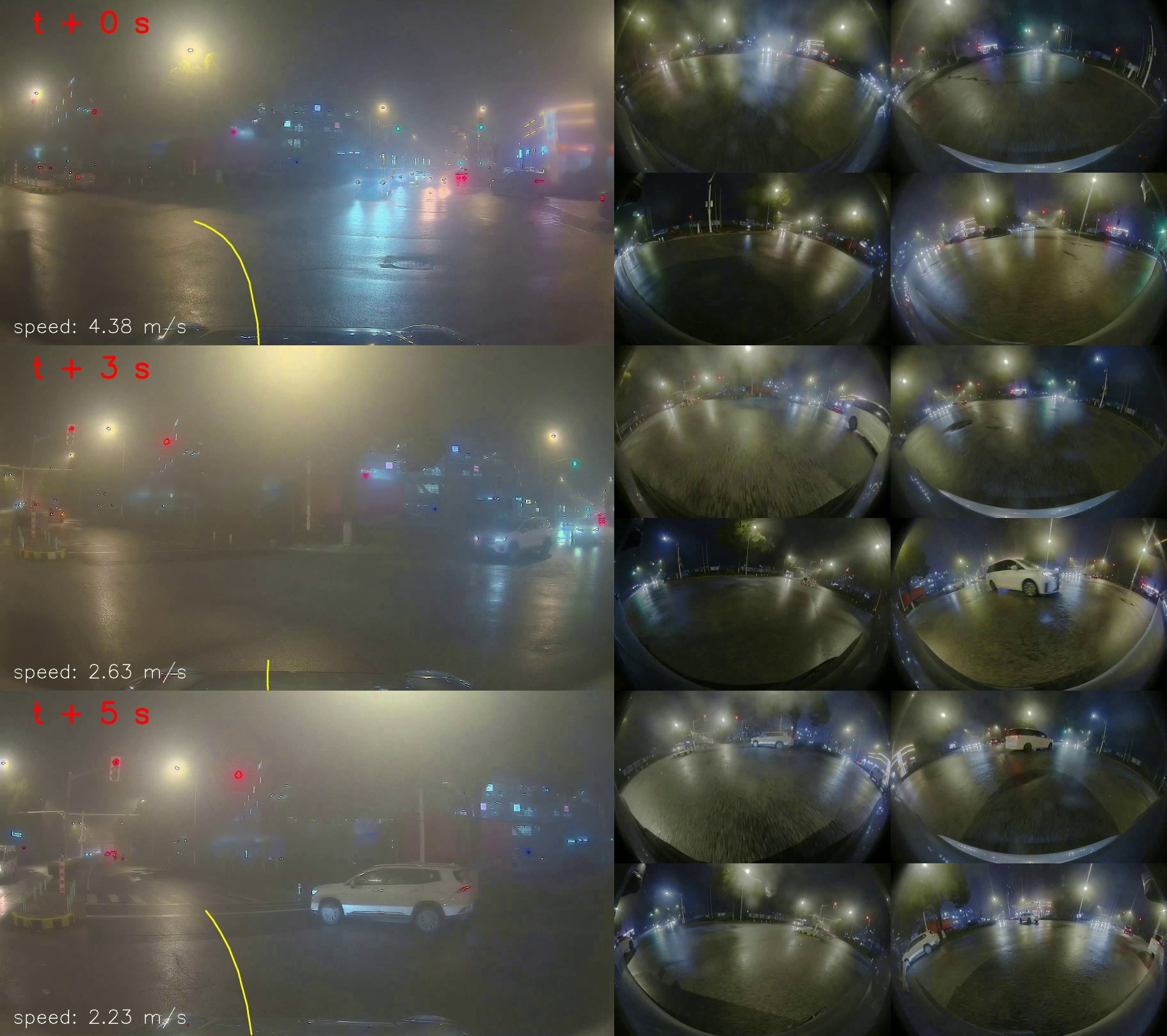}
    \caption{Unprotected left turn}
    \label{fig:test-B}
\end{subfigure}\\
\centering
\begin{subfigure}{0.48\linewidth}
  \centering
    \includegraphics[width=1\linewidth]{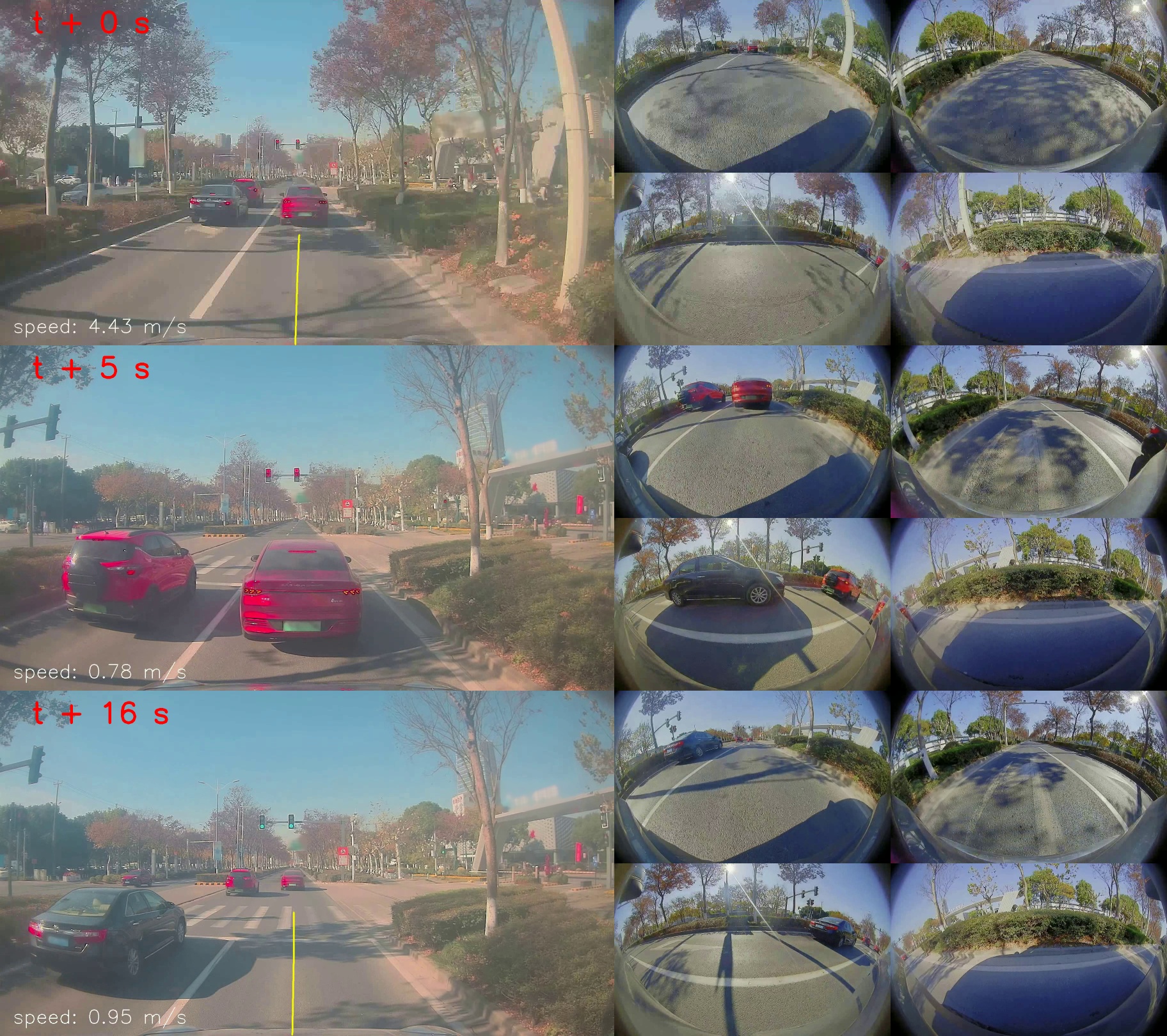}
    \caption{Reactions to traffic lights}
    \label{fig:test-C}
\end{subfigure}
\begin{subfigure}{0.48\linewidth}
  \centering
    \includegraphics[width=1\linewidth]{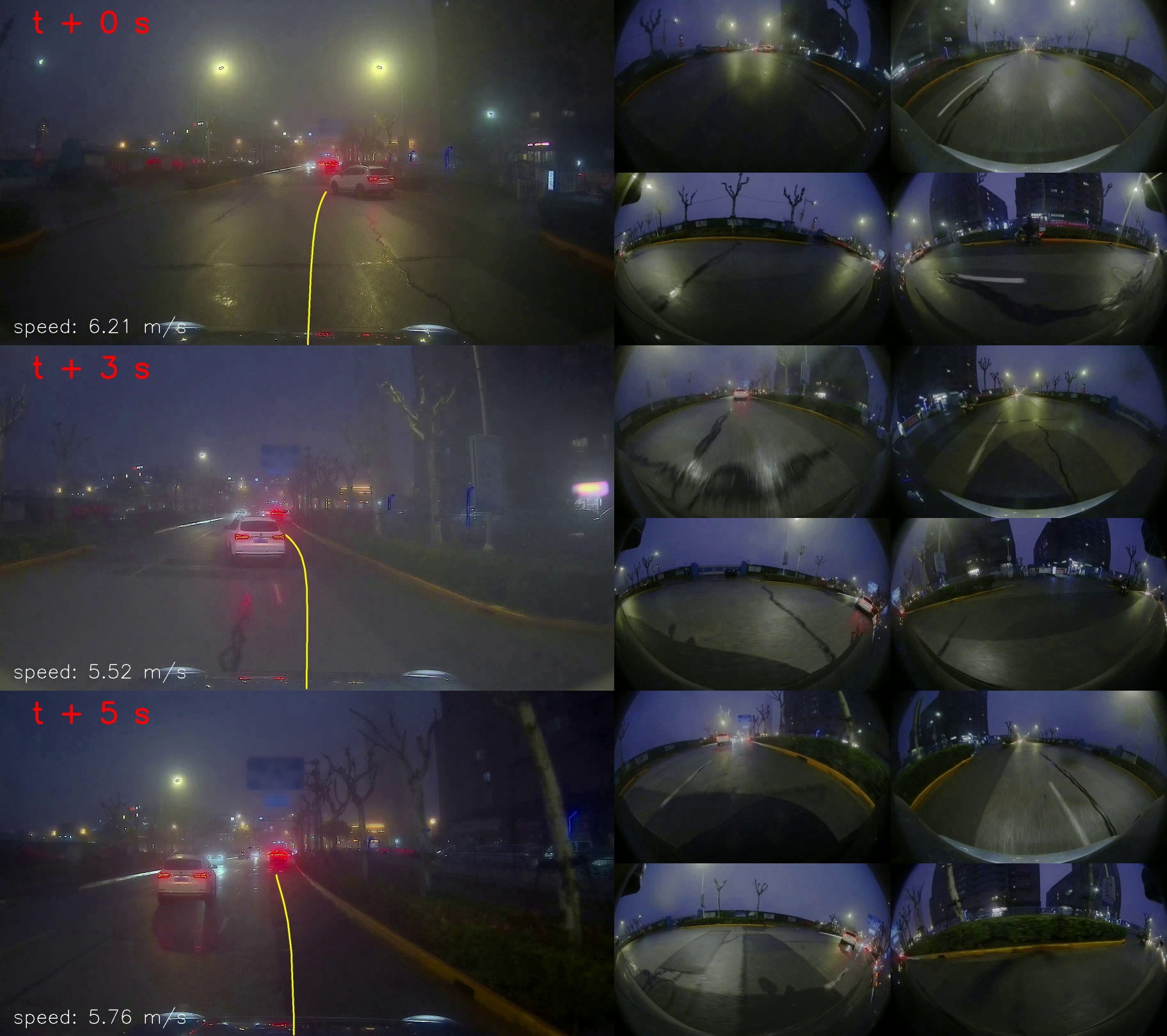}
    \caption{Avoid a moving vehicle}
    \label{fig:test-D}
\end{subfigure}
\caption{Real-world tests in urban areas. The front view image are combined with four surrounding view images and the output planning trajectories are projected as yellow lines. 
}
\label{fig:real_world_test}
\end{figure}

\section{Conclusion}
In this paper, we propose Int2Planner, an intention-based motion planner for integrated prediction and planning. We constrain the uncertainty of EA by route intention points, and multi-modal planning trajectories are generated and optimized based on route intention points.
The experimental results show that our model achieves the state-of-the-art performance. Route intention points effectively improve planning performance and provide reasonable intentions for EA. Furthermore, we deploy Int2Planner in real-world vehicles and the tests show that Int2Planner can continuously interact with the environment and output reasonable and safe planning trajectories. 
\noindent \textbf{Limitation and Future Work.} In closed-loop simulations and real-world vehicle tests, the output planning trajectory with the highest confidence is currently utilized, but this trajectory is not necessarily the optimal one, especially in complex traffic scenarios. 
In future work, we will focus on handling multi-modal planning to further improve performance.

\section*{Acknowledgements}
The work was in part supported by Shanghai Municipal Science and Technology Project under Grant 22Z510901584, 2021SHZDZX0102.

\bibliography{aaai25}

\begin{thebibliography}{41}
\providecommand{\natexlab}[1]{#1}

\bibitem[{Caesar et~al.(2021)Caesar, Kabzan, Tan, Fong, Wolff, Lang, Fletcher, Beijbom, and Omari}]{plancnn2021nuplan}
Caesar, H.; Kabzan, J.; Tan, K.~S.; Fong, W.~K.; Wolff, E.; Lang, A.; Fletcher, L.; Beijbom, O.; and Omari, S. 2021.
\newblock Nuplan: A closed-loop ML-based planning benchmark for autonomous vehicles.
\newblock \emph{arXiv preprint arXiv:2106.11810}.

\bibitem[{Casas, Luo, and Urtasun(2018)}]{casas2018intentnet}
Casas, S.; Luo, W.; and Urtasun, R. 2018.
\newblock Intentnet: Learning to predict intention from raw sensor data.
\newblock In \emph{Conference on Robot Learning}, 947--956. PMLR.

\bibitem[{Casas, Sadat, and Urtasun(2021)}]{Casas_2021_CVPR}
Casas, S.; Sadat, A.; and Urtasun, R. 2021.
\newblock MP3: A Unified Model To Map, Perceive, Predict and Plan.
\newblock In \emph{Proceedings of the IEEE/CVF Conference on Computer Vision and Pattern Recognition (CVPR)}, 14403--14412.

\bibitem[{Cheng et~al.(2023)Cheng, Chen, Mei, Yang, Li, and Liu}]{plantf2023}
Cheng, J.; Chen, Y.; Mei, X.; Yang, B.; Li, B.; and Liu, M. 2023.
\newblock Rethinking Imitation-based Planner for Autonomous Driving.
\newblock \emph{arXiv preprint arXiv:2309.10443}.

\bibitem[{Dauner et~al.(2023)Dauner, Hallgarten, Geiger, and Chitta}]{val142023parting}
Dauner, D.; Hallgarten, M.; Geiger, A.; and Chitta, K. 2023.
\newblock Parting with Misconceptions about Learning-based Vehicle Motion Planning.
\newblock \emph{arXiv preprint arXiv:2306.07962}.

\bibitem[{Deo, Wolff, and Beijbom(2022)}]{deo2022multimodal}
Deo, N.; Wolff, E.; and Beijbom, O. 2022.
\newblock Multimodal trajectory prediction conditioned on lane-graph traversals.
\newblock In \emph{Conference on Robot Learning}, 203--212. PMLR.

\bibitem[{Fang et~al.(2020)Fang, Jiang, Shi, and Zhou}]{fang2020tpnet}
Fang, L.; Jiang, Q.; Shi, J.; and Zhou, B. 2020.
\newblock Tpnet: Trajectory proposal network for motion prediction.
\newblock In \emph{Proceedings of the IEEE/CVF Conference on Computer Vision and Pattern Recognition}, 6797--6806.

\bibitem[{Gilles et~al.(2022)Gilles, Sabatini, Tsishkou, Stanciulescu, and Moutarde}]{gilles2022gohome}
Gilles, T.; Sabatini, S.; Tsishkou, D.; Stanciulescu, B.; and Moutarde, F. 2022.
\newblock GOHOME: Graph-oriented heatmap output for future motion estimation.
\newblock In \emph{2022 International Conference on Robotics and Automation (ICRA)}, 9107--9114.

\bibitem[{Gu, Sun, and Zhao(2021)}]{densetnt}
Gu, J.; Sun, C.; and Zhao, H. 2021.
\newblock DenseTNT: End-to-end Trajectory Prediction from Dense Goal Sets.
\newblock In \emph{2021 IEEE/CVF International Conference on Computer Vision (ICCV)}, 15283--15292.

\bibitem[{Hagedorn et~al.(2023)Hagedorn, Hallgarten, Stoll, and Condurache}]{review}
Hagedorn, S.; Hallgarten, M.; Stoll, M.; and Condurache, A. 2023.
\newblock Rethinking Integration of Prediction and Planning in Deep Learning-Based Automated Driving Systems: A Review.
\newblock \emph{arXiv preprint arXiv:2308.05731}.

\bibitem[{Hallgarten, Stoll, and Zell(2023)}]{gc-pgp2023prediction}
Hallgarten, M.; Stoll, M.; and Zell, A. 2023.
\newblock From Prediction to Planning With Goal Conditioned Lane Graph Traversals.
\newblock \emph{arXiv preprint arXiv:2302.07753}.

\bibitem[{Hu et~al.(2017)Hu, Liao, Wang, Li, Cheng, and Chen}]{hu2017decision}
Hu, M.; Liao, Y.; Wang, W.; Li, G.; Cheng, B.; and Chen, F. 2017.
\newblock Decision tree-based maneuver prediction for driver rear-end risk-avoidance behaviors in cut-in scenarios.
\newblock \emph{Journal of Advanced Transportation}, 2017.

\bibitem[{Hu et~al.(2023{\natexlab{a}})Hu, Yang, Chen, Li, Sima, Zhu, Chai, Du, Lin, Wang, Lu, Jia, Liu, Dai, Qiao, and Li}]{plan1}
Hu, Y.; Yang, J.; Chen, L.; Li, K.; Sima, C.; Zhu, X.; Chai, S.; Du, S.; Lin, T.; Wang, W.; Lu, L.; Jia, X.; Liu, Q.; Dai, J.; Qiao, Y.; and Li, H. 2023{\natexlab{a}}.
\newblock Planning-Oriented Autonomous Driving.
\newblock In \emph{Proceedings of the IEEE/CVF Conference on Computer Vision and Pattern Recognition (CVPR)}, 17853--17862.

\bibitem[{Hu et~al.(2023{\natexlab{b}})Hu, Yang, Chen, Li, Sima, Zhu, Chai, Du, Lin, Wang, Lu, Jia, Liu, Dai, Qiao, and Li}]{uniad_Hu_2023_CVPR}
Hu, Y.; Yang, J.; Chen, L.; Li, K.; Sima, C.; Zhu, X.; Chai, S.; Du, S.; Lin, T.; Wang, W.; Lu, L.; Jia, X.; Liu, Q.; Dai, J.; Qiao, Y.; and Li, H. 2023{\natexlab{b}}.
\newblock Planning-Oriented Autonomous Driving.
\newblock In \emph{Proceedings of the IEEE/CVF Conference on Computer Vision and Pattern Recognition (CVPR)}, 17853--17862.

\bibitem[{Huang, Liu, and Lv(2023)}]{huang2023gameformer}
Huang, Z.; Liu, H.; and Lv, C. 2023.
\newblock GameFormer: Game-theoretic Modeling and Learning of Transformer-based Interactive Prediction and Planning for Autonomous Driving.
\newblock \emph{arXiv preprint arXiv:2303.05760}.

\bibitem[{Jia et~al.(2023{\natexlab{a}})Jia, Gao, Chen, Yan, Liu, and Li}]{plan2}
Jia, X.; Gao, Y.; Chen, L.; Yan, J.; Liu, P.~L.; and Li, H. 2023{\natexlab{a}}.
\newblock DriveAdapter: Breaking the Coupling Barrier of Perception and Planning in End-to-End Autonomous Driving.
\newblock In \emph{Proceedings of the IEEE/CVF International Conference on Computer Vision (ICCV)}, 7953--7963.

\bibitem[{Jia et~al.(2023{\natexlab{b}})Jia, Wu, Chen, Liu, Li, and Yan}]{jia2023hdgt}
Jia, X.; Wu, P.; Chen, L.; Liu, Y.; Li, H.; and Yan, J. 2023{\natexlab{b}}.
\newblock Hdgt: Heterogeneous driving graph transformer for multi-agent trajectory prediction via scene encoding.
\newblock \emph{IEEE Transactions on Pattern Analysis and Machine Intelligence}.

\bibitem[{Jiang et~al.(2023)Jiang, Chen, Xu, Liao, Chen, Zhou, Zhang, Liu, Huang, and Wang}]{jiang2023vad}
Jiang, B.; Chen, S.; Xu, Q.; Liao, B.; Chen, J.; Zhou, H.; Zhang, Q.; Liu, W.; Huang, C.; and Wang, X. 2023.
\newblock VAD: Vectorized Scene Representation for Efficient Autonomous Driving.
\newblock \emph{ICCV}.

\bibitem[{Li et~al.(2017)Li, Dai, Li, Li, and Di}]{li2017real}
Li, J.; Dai, B.; Li, X.; Li, C.; and Di, Y. 2017.
\newblock A real-time and predictive trajectory-generation motion planner for autonomous ground vehicles.
\newblock In \emph{2017 9th International Conference on Intelligent Human-Machine Systems and Cybernetics (IHMSC)}, volume~2, 108--113.

\bibitem[{Li et~al.(2021)Li, Eiffert, Shan, Gomez-Donoso, Worrall, and Nebot}]{li2021attentional}
Li, K.; Eiffert, S.; Shan, M.; Gomez-Donoso, F.; Worrall, S.; and Nebot, E. 2021.
\newblock Attentional-GCNN: Adaptive pedestrian trajectory prediction towards generic autonomous vehicle use cases.
\newblock In \emph{2021 IEEE International Conference on Robotics and Automation (ICRA)}, 14241--14247.

\bibitem[{Lin et~al.(2022)Lin, Li, Bi, and Qin}]{9349962}
Lin, L.; Li, W.; Bi, H.; and Qin, L. 2022.
\newblock Vehicle Trajectory Prediction Using LSTMs With Spatial–Temporal Attention Mechanisms.
\newblock \emph{IEEE Intelligent Transportation Systems Magazine}, 14(2): 197--208.

\bibitem[{Liu et~al.(2021)Liu, Zhang, Fang, Jiang, and Zhou}]{liu2021multimodal}
Liu, Y.; Zhang, J.; Fang, L.; Jiang, Q.; and Zhou, B. 2021.
\newblock Multimodal motion prediction with stacked transformers.
\newblock In \emph{Proceedings of the IEEE/CVF conference on computer vision and pattern recognition}, 7577--7586.

\bibitem[{Mo et~al.(2022)Mo, Huang, Xing, and Lv}]{9700483}
Mo, X.; Huang, Z.; Xing, Y.; and Lv, C. 2022.
\newblock Multi-Agent Trajectory Prediction With Heterogeneous Edge-Enhanced Graph Attention Network.
\newblock \emph{IEEE Transactions on Intelligent Transportation Systems}, 23(7): 9554--9567.

\bibitem[{Nayakanti et~al.(2023)Nayakanti, Al-Rfou, Zhou, Goel, Refaat, and Sapp}]{nayakanti2023wayformer}
Nayakanti, N.; Al-Rfou, R.; Zhou, A.; Goel, K.; Refaat, K.~S.; and Sapp, B. 2023.
\newblock Wayformer: Motion forecasting via simple \& efficient attention networks.
\newblock In \emph{2023 IEEE International Conference on Robotics and Automation (ICRA)}, 2980--2987.

\bibitem[{Ngiam et~al.(2022)Ngiam, Vasudevan, Caine, Zhang, Chiang, Ling, Roelofs, Bewley, Liu, Venugopal, Weiss, Sapp, Chen, and Shlens}]{ngiam2021scene}
Ngiam, J.; Vasudevan, V.; Caine, B.; Zhang, Z.; Chiang, H.-T.~L.; Ling, J.; Roelofs, R.; Bewley, A.; Liu, C.; Venugopal, A.; Weiss, D.~J.; Sapp, B.; Chen, Z.; and Shlens, J. 2022.
\newblock Scene Transformer: A unified architecture for predicting future trajectories of multiple agents.
\newblock In \emph{International Conference on Learning Representations (ICLR)}.

\bibitem[{Nilsson et~al.(2013)Nilsson, Ali, Falcone, and Sj{\"o}berg}]{nilsson2013predictive}
Nilsson, J.; Ali, M.; Falcone, P.; and Sj{\"o}berg, J. 2013.
\newblock Predictive manoeuvre generation for automated driving.
\newblock In \emph{16th International IEEE Conference on Intelligent Transportation Systems (ITSC)}, 418--423.

\bibitem[{Renz et~al.(2022)Renz, Chitta, Mercea, Koepke, Akata, and Geiger}]{renz2022plant}
Renz, K.; Chitta, K.; Mercea, O.-B.; Koepke, A.; Akata, Z.; and Geiger, A. 2022.
\newblock Plant: Explainable planning transformers via object-level representations.
\newblock \emph{arXiv preprint arXiv:2210.14222}.

\bibitem[{Rhinehart et~al.(2019)Rhinehart, McAllister, Kitani, and Levine}]{rhinehart2019precog}
Rhinehart, N.; McAllister, R.; Kitani, K.; and Levine, S. 2019.
\newblock Precog: Prediction conditioned on goals in visual multi-agent settings.
\newblock In \emph{Proceedings of the IEEE/CVF International Conference on Computer Vision}, 2821--2830.

\bibitem[{Sadat et~al.(2020)Sadat, Casas, Ren, Wu, Dhawan, and Urtasun}]{sadat2020perceive}
Sadat, A.; Casas, S.; Ren, M.; Wu, X.; Dhawan, P.; and Urtasun, R. 2020.
\newblock Perceive, predict, and plan: Safe motion planning through interpretable semantic representations.
\newblock In \emph{Proceedings of the European Conference on Computer Vision (ECCV)}, 414--430. Springer.

\bibitem[{Schmerling et~al.(2018)Schmerling, Leung, Vollprecht, and Pavone}]{schmerling2018multimodal}
Schmerling, E.; Leung, K.; Vollprecht, W.; and Pavone, M. 2018.
\newblock Multimodal probabilistic model-based planning for human-robot interaction.
\newblock In \emph{2018 IEEE International Conference on Robotics and Automation (ICRA)}, 3399--3406.

\bibitem[{Shi et~al.(2022)Shi, Jiang, Dai, and Schiele}]{shi2022mtr}
Shi, S.; Jiang, L.; Dai, D.; and Schiele, B. 2022.
\newblock Motion transformer with global intention localization and local movement refinement.
\newblock \emph{Advances in Neural Information Processing Systems}, 35: 6531--6543.

\bibitem[{Sun et~al.(2022)Sun, Huang, Gu, Williams, and Zhao}]{sun2022m2i}
Sun, Q.; Huang, X.; Gu, J.; Williams, B.; and Zhao, H. 2022.
\newblock M2i: From factored marginal trajectory prediction to interactive prediction.
\newblock In \emph{Proceedings of the IEEE/CVF Conference on Computer Vision and Pattern Recognition (CVPR)}, 6543--6552.

\bibitem[{Treiber, Hennecke, and Helbing(2000)}]{idm2000congested}
Treiber, M.; Hennecke, A.; and Helbing, D. 2000.
\newblock Congested traffic states in empirical observations and microscopic simulations.
\newblock \emph{Physical review E}, 62(2): 1805.

\bibitem[{Van~Hoek, Ploeg, and Nijmeijer(2021)}]{9415170}
Van~Hoek, R.; Ploeg, J.; and Nijmeijer, H. 2021.
\newblock Cooperative Driving of Automated Vehicles Using B-Splines for Trajectory Planning.
\newblock \emph{IEEE Transactions on Intelligent Vehicles}, 6(3): 594--604.

\bibitem[{Varadarajan et~al.(2022)Varadarajan, Hefny, Srivastava, Refaat, Nayakanti, Cornman, Chen, Douillard, Lam, Anguelov, and Sapp}]{varadarajan2022multipath++}
Varadarajan, B.; Hefny, A.; Srivastava, A.; Refaat, K.~S.; Nayakanti, N.; Cornman, A.; Chen, K.; Douillard, B.; Lam, C.~P.; Anguelov, D.; and Sapp, B. 2022.
\newblock Multipath++: Efficient information fusion and trajectory aggregation for behavior prediction.
\newblock In \emph{2022 International Conference on Robotics and Automation (ICRA)}, 7814--7821.

\bibitem[{Xie et~al.(2018)Xie, Gao, Huang, Qian, and Wang}]{xie2018driving}
Xie, G.; Gao, H.; Huang, B.; Qian, L.; and Wang, J. 2018.
\newblock A driving behavior awareness model based on a dynamic Bayesian network and distributed genetic algorithm.
\newblock \emph{International Journal of Computational Intelligence Systems}, 11(1): 469--482.

\bibitem[{Xie et~al.(2017)Xie, Gao, Qian, Huang, Li, and Wang}]{xie2017vehicle}
Xie, G.; Gao, H.; Qian, L.; Huang, B.; Li, K.; and Wang, J. 2017.
\newblock Vehicle trajectory prediction by integrating physics-and maneuver-based approaches using interactive multiple models.
\newblock \emph{IEEE Transactions on Industrial Electronics}, 65(7): 5999--6008.

\bibitem[{Ye, Cao, and Chen(2021)}]{ye2021tpcn}
Ye, M.; Cao, T.; and Chen, Q. 2021.
\newblock TPCN: Temporal point cloud networks for motion forecasting.
\newblock In \emph{Proceedings of the IEEE/CVF Conference on Computer Vision and Pattern Recognition (CVPR)}, 11318--11327.

\bibitem[{Ye et~al.(2023)Ye, Jing, Hu, Huang, Gao, Li, Wang, Guo, Xiao, Mao, Zheng, Li, Chen, and Yu}]{ye2023fusionad}
Ye, T.; Jing, W.; Hu, C.; Huang, S.; Gao, L.; Li, F.; Wang, J.; Guo, K.; Xiao, W.; Mao, W.; Zheng, H.; Li, K.; Chen, J.; and Yu, K. 2023.
\newblock Fusionad: Multi-modality fusion for prediction and planning tasks of autonomous driving.
\newblock \emph{arXiv preprint arXiv:2308.01006}.

\bibitem[{Zhao et~al.(2021)Zhao, Gao, Lan, Sun, Sapp, Varadarajan, Shen, Shen, Chai, Schmid, Li, and Anguelov}]{zhao2021tnt}
Zhao, H.; Gao, J.; Lan, T.; Sun, C.; Sapp, B.; Varadarajan, B.; Shen, Y.; Shen, Y.; Chai, Y.; Schmid, C.; Li, C.; and Anguelov, D. 2021.
\newblock TNT: Target-driven Trajectory Prediction.
\newblock In \emph{Proceedings of the 2020 Conference on Robot Learning}, volume 155, 895--904.

\bibitem[{Zhou et~al.(2023)Zhou, Wang, Li, and Huang}]{zhou2023query}
Zhou, Z.; Wang, J.; Li, Y.-H.; and Huang, Y.-K. 2023.
\newblock Query-centric trajectory prediction.
\newblock In \emph{Proceedings of the IEEE/CVF Conference on Computer Vision and Pattern Recognition}, 17863--17873.

\end{thebibliography}

\newpage

\section*{Appendix}
\appendix

\setcounter{figure}{0}
\setcounter{table}{0}
\renewcommand{\thefigure}{A\arabic{figure}}
\renewcommand{\thetable}{A\arabic{table}}

\subsection{Model design}
Table~\ref{tab:model_detail} shows the modeling details compared with other planners. Int2Planner is designed as a purely learn-based planner, which can generate planning trajectory without combining ruled-based methods. However, rule-based post-processing method can also be implemented to optimize the output planning trajectory. The rule-based post-processing mainly considers the safety issues of planning. It uses route paths to constrain the overall direction of the planned trajectory, and utilize the integrated prediction trajectories to perform collision check. Based on this, the details of the planning trajectory are optimized. 

\begin{table}[th!]
\footnotesize
\begin{center}

\resizebox{0.46\textwidth}{!}{
\begin{tabular}{@{} lccc}
\toprule 
Model & Integrated & Route Conditioning & Rule/Learn-based \\
\midrule
IDM & $\times$ & Input Features & Rule-base \\
PlanCNN & $\times$ & Input Features & Learn-based \\
GC-PGP & $\times$ & Route Attention & Learn-based \\
PDM-Hybird & $\times$ & Route Attention & Hybrid \\
PlanTF & \checkmark & Input Features & Learn-based \\
GameFormer & \checkmark & Route Attention & Hybrid \\
Int2Planner (ours) & \checkmark & \makecell{Route Attention\\Route Intention} & Learn-base \\
\bottomrule 
\end{tabular}
}
\caption{Comparison of modeling details with state-of-the-art methods.} 
\label{tab:model_detail}
\end{center}
\end{table}

\subsection{Private Dataset}
\subsubsection{Object Types}
\begin{figure}[th!]
  \centering
  \includegraphics[width=0.95\linewidth]{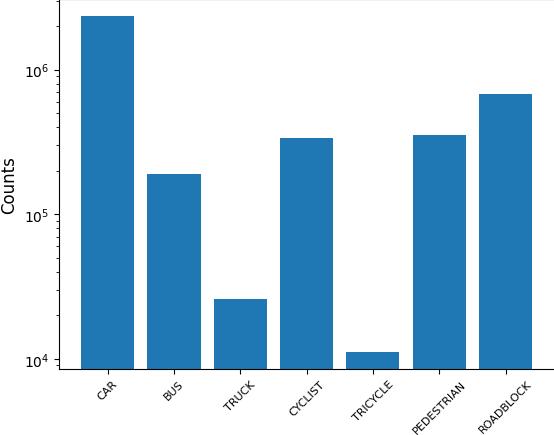}
   \caption{The object type distribution in the private dataset.}
   \label{fig:objec_count}
\end{figure}

The objects in the private dataset are labeled by a state-of-the-art offline perception system, mainly including following object types: "CAR", "BUS", "TRUCK", "CYCLIST", "TRICYCLE", "PEDESTRIAN" and "ROADBLOCK". The distribution of object types are shown in Fig.~\ref{fig:objec_count}, which is calculated from 100,000 scenes of the private dataset.

\subsubsection{Scenes} 
\begin{figure*}[t!]
  \centering
\begin{subfigure}{0.33\linewidth}
  \centering
    \includegraphics[width=1\linewidth]{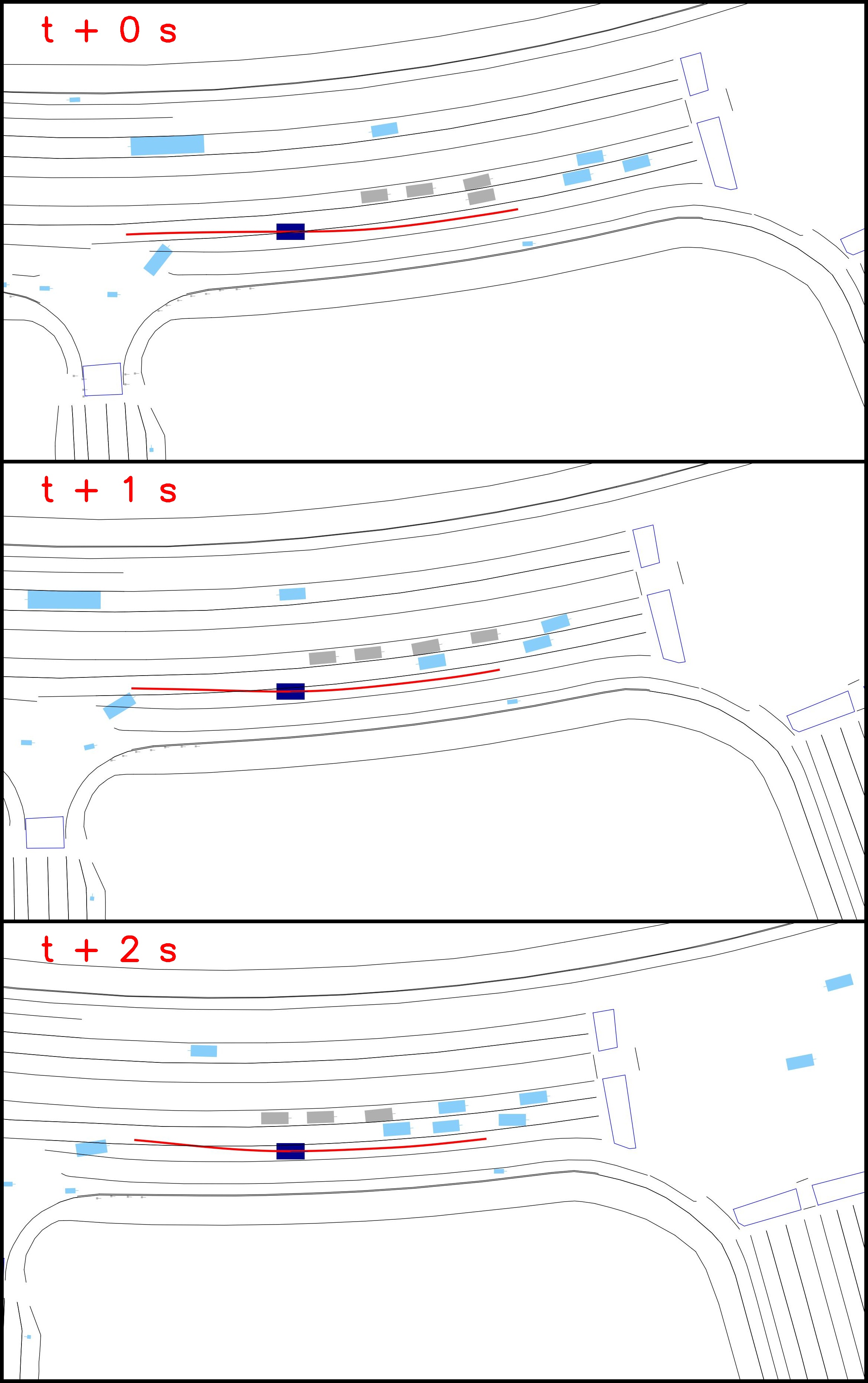}
    \caption{Lane change}
    \label{fig:scene-a}
\end{subfigure}
\begin{subfigure}{0.33\linewidth}
  \centering
    \includegraphics[width=1\linewidth]{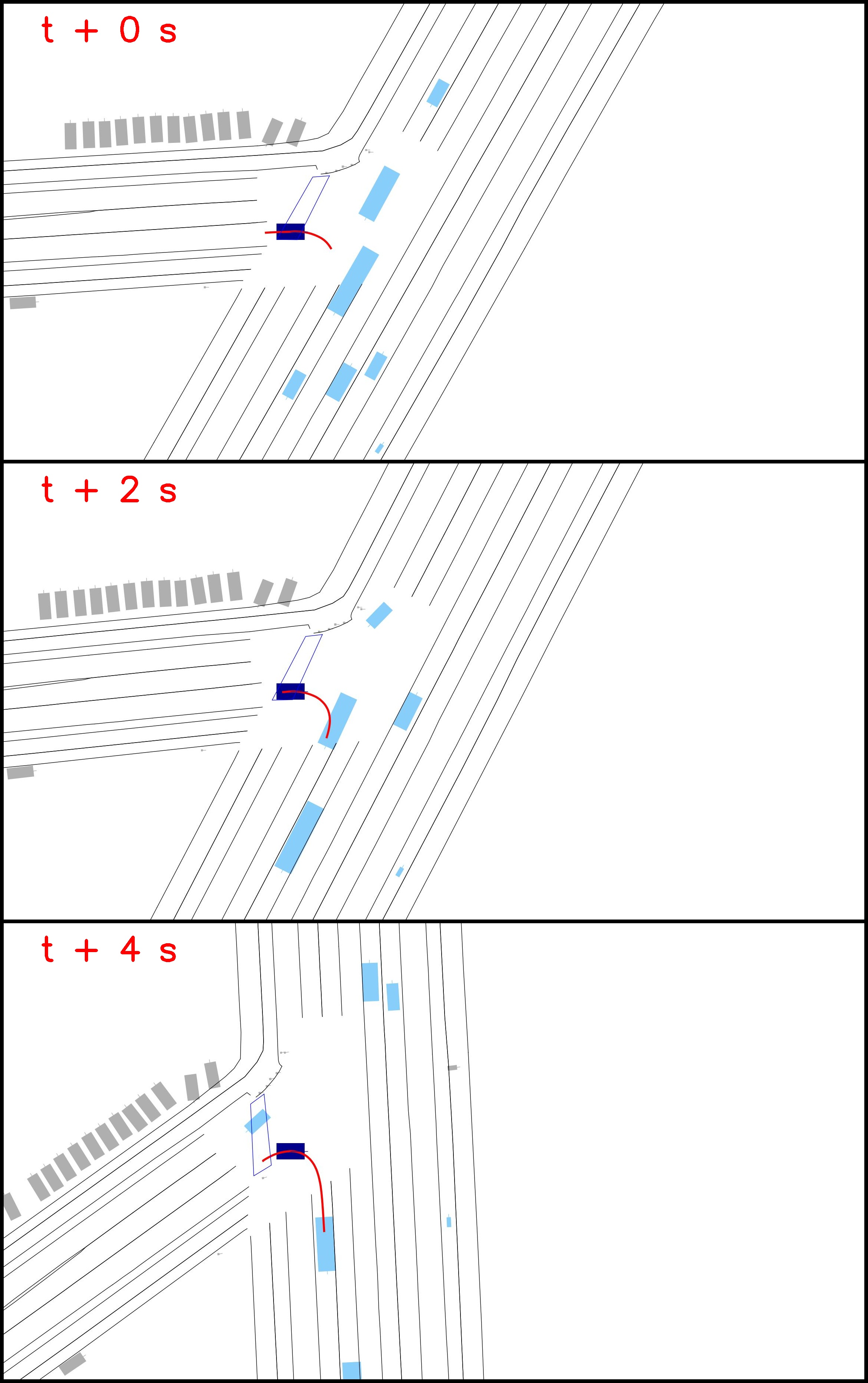}
    \caption{Sharp angle turn}
    \label{fig:scene-b}
\end{subfigure}
\centering
\begin{subfigure}{0.33\linewidth}
  \centering
    \includegraphics[width=1\linewidth]{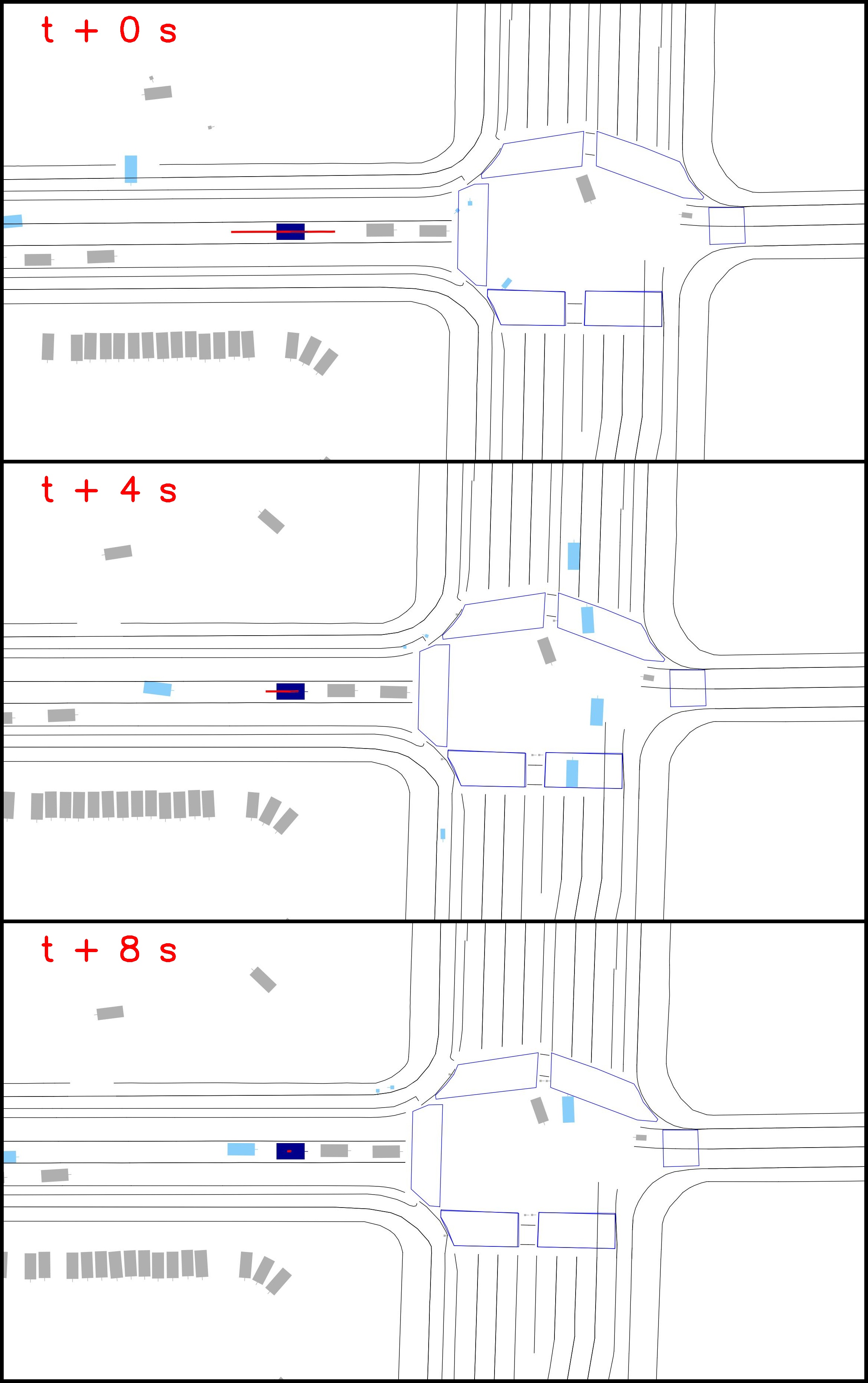}
    \caption{Stop with lead}
    \label{fig:scene-c}
\end{subfigure}\\
\centering
\begin{subfigure}{0.33\linewidth}
  \centering
    \includegraphics[width=1\linewidth]{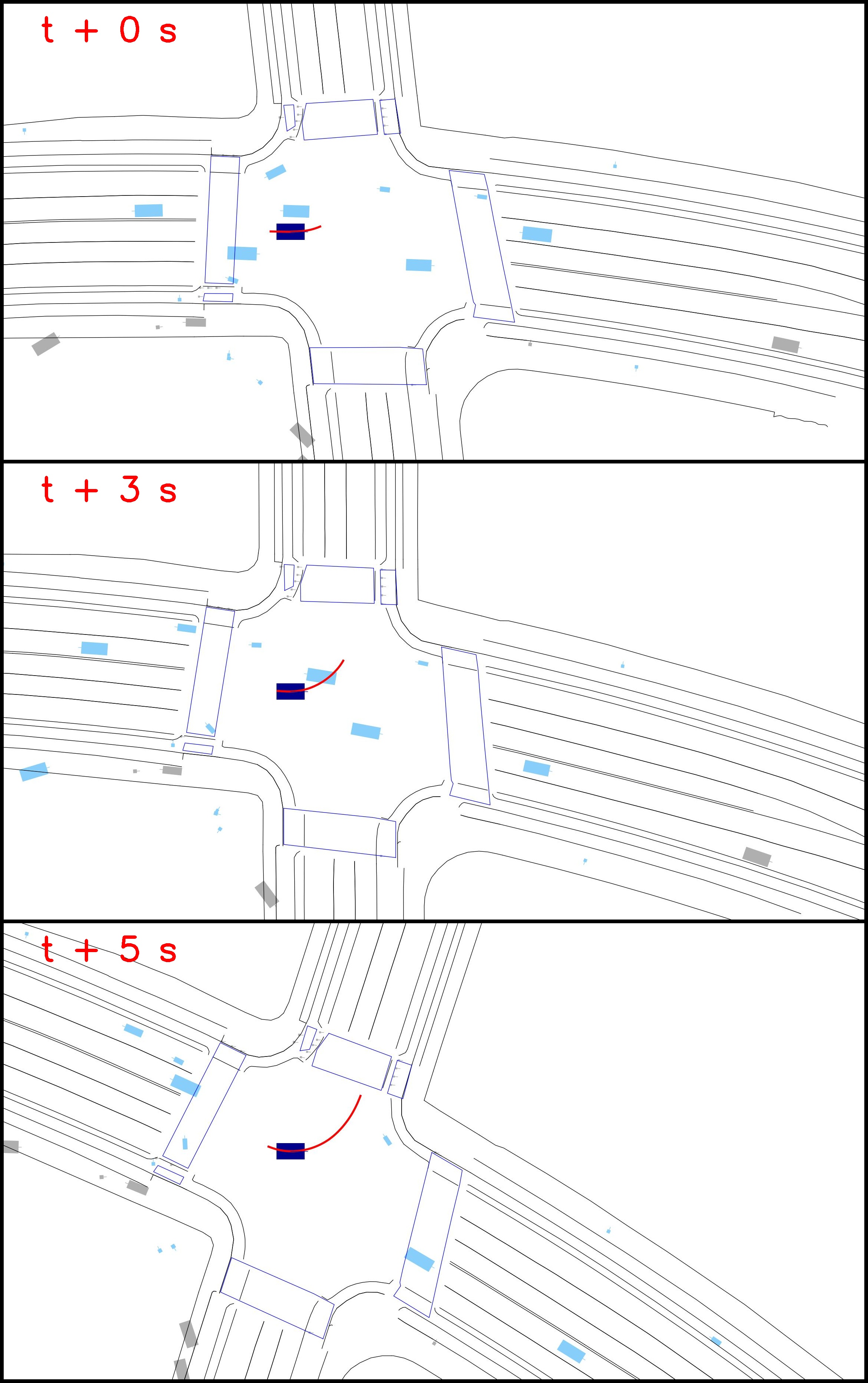}
    \caption{Unprotected left turn}
    \label{fig:scene-d}
\end{subfigure}
\begin{subfigure}{0.33\linewidth}
  \centering
    \includegraphics[width=1\linewidth]{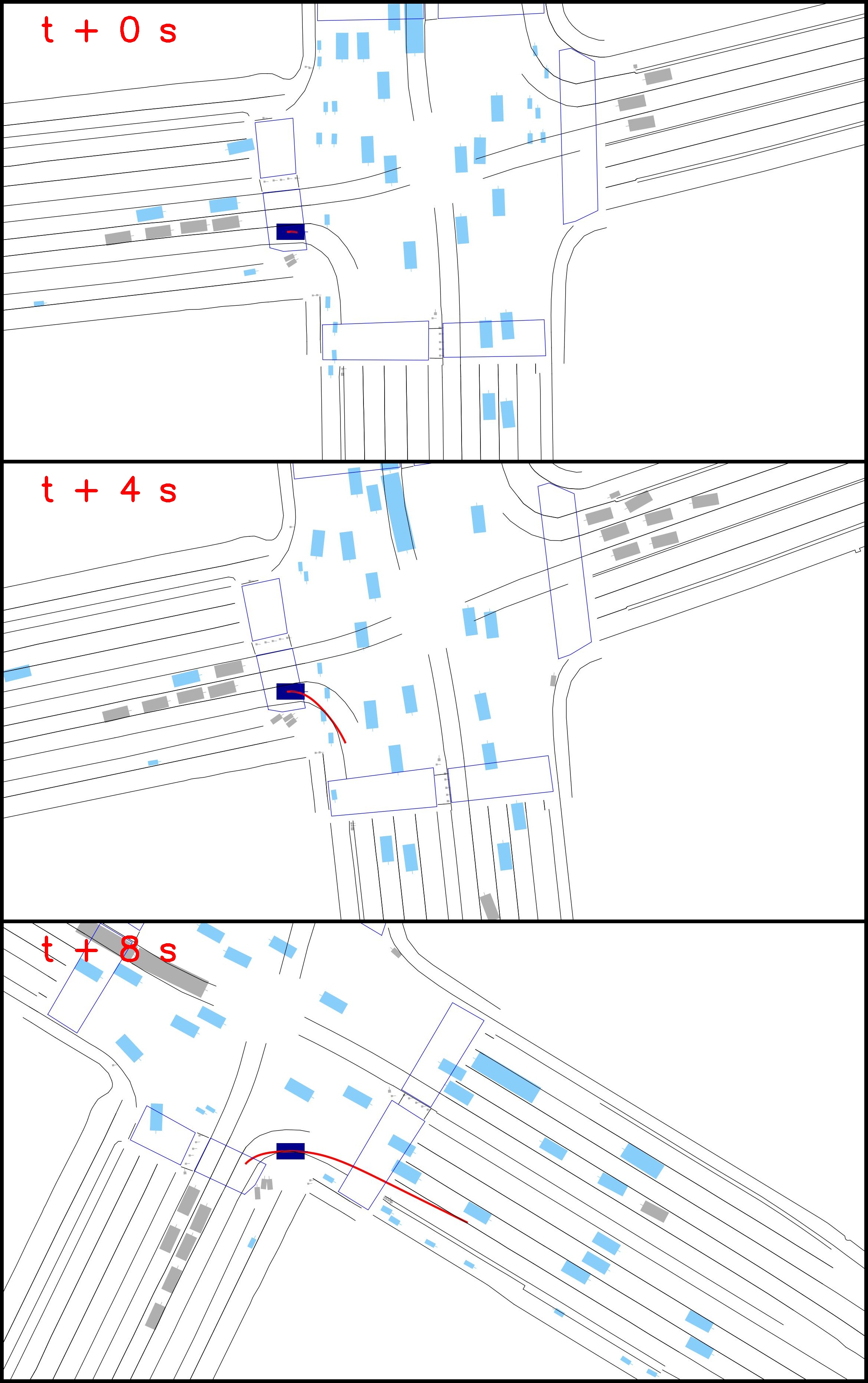}
    \caption{Interact with cyclists}
    \label{fig:scene-e}
\end{subfigure}
\begin{subfigure}{0.33\linewidth}
  \centering
    \includegraphics[width=1\linewidth]{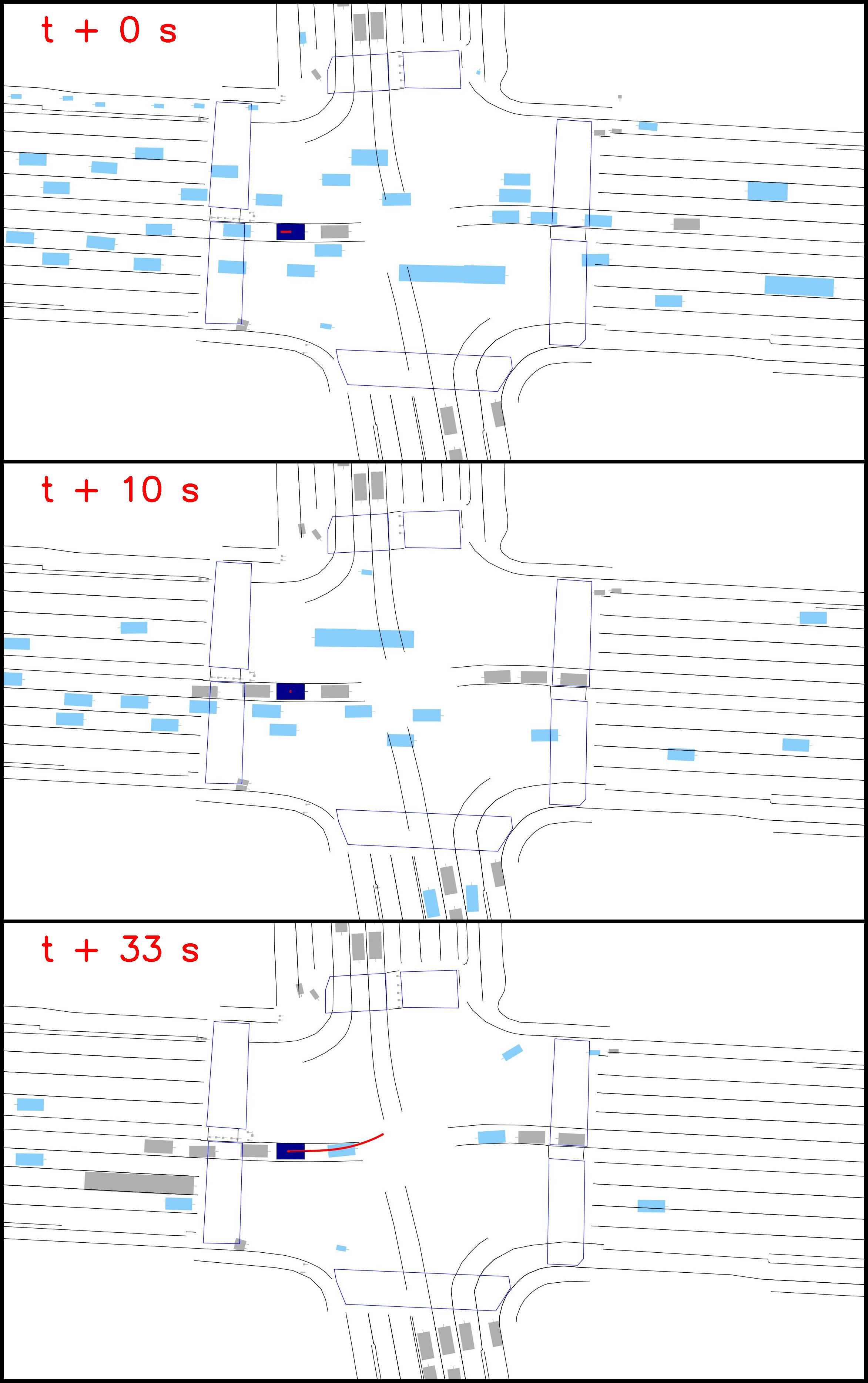}
    \caption{Stationary in traffic}
    \label{fig:scene-f}
\end{subfigure}

\caption{Visualization of typical scenes from the private dataset. The bounding boxes of EA, moving SA and static SA are colored in deep blue, light blue and grey, respectively. The trajectory of EA is plotted in red lines. Lane divides and crosswalks are also displayed.
}
\label{fig:scenes}
\end{figure*}

Fig.~\ref{fig:scenes} 
demonstrates several typical scenes from the private dataset, including lane change, sharp angle turn, stop/following with lead, interaction with cyclists and pedestrians, left and right turn, moving/stationary in traffic and nearby long vehicles.

\subsection{Additional Experiment Results}

\subsubsection{Evaluations on Nuplan}Table~\ref{tab:nuplan} shows the experimental results of nuPlan dataset, which are evaluated on the aforementioned Val14 benchmark. Compared with GameFormer planner, Int2Planner demonstrates better planning performance, and also has better prediction FDE metric. Compared with PlanTF, both trained with 1M scenarios, Int2Planner achieves better plan ADE and FDE. These results reveal the superiority of Int2Planner in planning and prediction tasks.

\subsubsection{Post-Processing}
To enhance the output of Int2Planner in closed-loop simulations, we add trajectory-based post-processing on top of the purely learning-based model. First, a reference line is generated from the route path to constrain the lateral displacement of the planning trajectory. Then, collision detection is performed between the prediction trajectory and the planning trajectory. The final planned trajectory is generated through iterative optimization based on these constrains.
\begin{table*}[tb!]

\begin{center}

\begin{tabular}{@{} lccccc @{} }
\toprule
\multirow{2}{*}{Model} & \multirow{2}{*}{$\#$Scenario} & \multicolumn{2}{c}{Plan} & \multicolumn{2}{c}{Prediction} \\
\cmidrule{3-6}
& & $\operatorname{ADE~}(\downarrow)$ & $\operatorname{FDE~}(\downarrow)$ & $\operatorname{minADE_6~}(\downarrow)$ & $\operatorname{minFDE_6~}(\downarrow)$ \\
\midrule 
GameFormer Planner & \multirow{3}{*}{400K} & 1.6497 & 4.3743 & \textbf{0.5294} & 1.3750\\
Int2Planner (ours, w/o Pred.) &  & 1.5460 & 4.3762 & - & -  \\ 
Int2Planner (ours) &  & \textbf{1.5247} & \textbf{4.2201} & 0.5328 & \textbf{1.1113}  \\ 
\midrule
PlanTF & \multirow{3}{*}{1M} & 1.6811 & 4.2058 & - & - \\ 
Int2Planner (ours, w/o Pred.) &  & 1.4848 & 4.1040 & - & -  \\ 
Int2Planner (ours) & & \textbf{1.4352} & \textbf{3.9622} & 0.5136 & 1.0570 \\
\bottomrule
\end{tabular}
\caption{Comparison of evaluation on the Val14 benchmark. 
"$\#$Scenario" denotes the number of scenarios used in the train set.
\textbf{Effectiveness of integrated prediction.} "w/o Pred." denotes that the prediction task is removed from Int2Planner.
} 
\label{tab:nuplan}
\end{center}
\end{table*}

\begin{figure*}[t!]
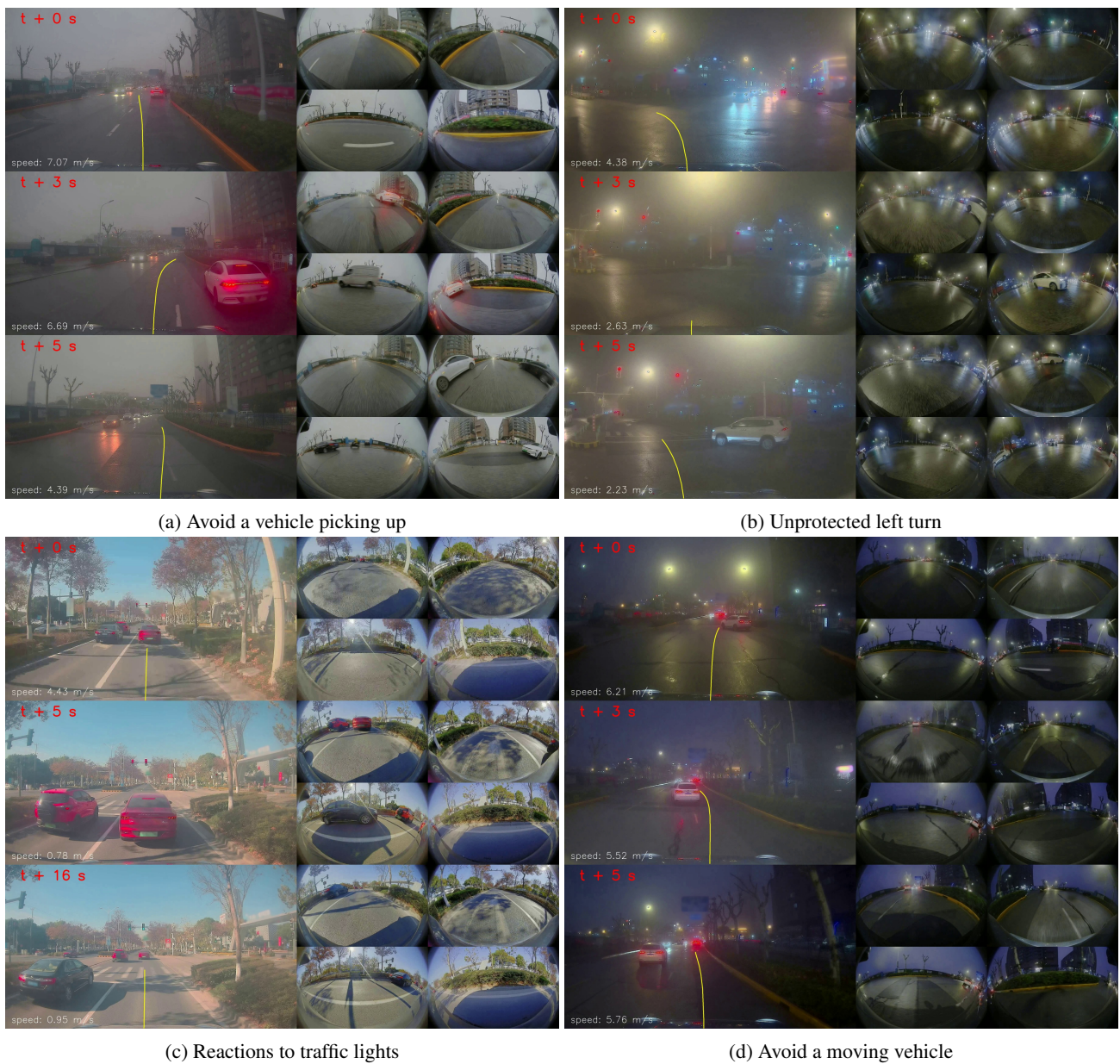

  \centering
\begin{subfigure}{0.48\linewidth}
  \centering
    \includegraphics[width=1\linewidth]{image/clip_pickup_0.5.jpg}
    \caption{Avoid a vehicle picking up}
    \label{fig:test-a}
\end{subfigure}
\begin{subfigure}{0.48\linewidth}
  \centering
    \includegraphics[width=1\linewidth]{image/clip_left_turn_0.5.jpg}
    \caption{Unprotected left turn}
    \label{fig:test-b}
\end{subfigure}\\
\centering
\begin{subfigure}{0.48\linewidth}
  \centering
    \includegraphics[width=1\linewidth]{image/clip_tl_1_0.5.jpg}
    \caption{Reactions to traffic lights}
    \label{fig:test-c}
\end{subfigure}
\begin{subfigure}{0.48\linewidth}
  \centering
    \includegraphics[width=1\linewidth]{image/clip_avoid_0.5.jpg}
    \caption{Avoid a moving vehicle}
    \label{fig:test-d}
\end{subfigure}
\caption{Real-world tests in urban areas. The front view image are combined with four surrounding view images and the output planning trajectories are projected as yellow lines onto the front view image.
}
\label{fig:test}
\end{figure*}

\subsubsection{Real-world Vehicle Test Analyze}
Fig.~\ref{fig:test} in the main content shows several real-world test scenarios and the corresponding videos of these scenarios are submitted as supplementary material.
Fig.~\ref{fig:test-a} shows that EA overtakes a vehicle, which has stopped to pick up passengers on the road size. Fig.~\ref{fig:test-b} shows that EA is making an unprotected left turn. It interacts with other vehicles, slows down to give way to a vehicle turning right and finally completes the left turn. Fig.~\ref{fig:test-c} demonstrates that EA stops behind the front vehicle at an intersection when the traffic light is red, and restarts after traffic light turns green. Fig.~\ref{fig:test-d} demonstrates When a slow-moving vehicle exits from a branch road, EA slows down and avoids the vehicle by changing lane to the left. It is worth noting that EA waits for another vehicle on the rear right to pass through before changing to the right neighbor lane.

It should be noted that in these real-world vehicle tests, we directly used Int2planner's planning trajectory to control the vehicle's motion without adding any rule-based post-processing, such as collision checking, drivable area compliance, etc. Therefore, abnormal vehicle behavior due to model failure may still occur during the testing process. In order to ensure the safety of real-world vehicle testing, we have equipped experienced on-board operators throughout the entire process of autonomous driving testing to take over dangerous driving situations. At the same time, the vehicles are also connected to a remote monitoring system to further ensure the safety of the testing process. Therefore, before applying Int2Planner to actual autonomous vehicles, users should be aware of the potential defects and corresponding negative effects of the model, and similar safety measures should also be taken during real-world testing and other applications.

\end{document}